\documentclass[pdflatex,sn-vancouver-num]{sn-jnl}%

\usepackage{graphicx}%
\usepackage{multirow}%
\usepackage{amsmath,amssymb,amsfonts}%
\usepackage{amsthm}%
\usepackage{mathrsfs}%
\usepackage[title]{appendix}%
\usepackage{xcolor}%
\usepackage{textcomp}%
\usepackage{manyfoot}%
\usepackage{booktabs}%
\usepackage{algorithm}%
\usepackage{algorithmicx}%
\usepackage{algpseudocode}%
\usepackage{listings}%
\usepackage{rotating}
\usepackage{mdframed}
\usepackage{caption}
\usepackage{verbatim}
\usepackage{placeins}

\theoremstyle{thmstyleone}%
\theoremstyle{thmstyletwo}%

\theoremstyle{thmstylethree}%

\begin{document}

\title[Article Title]{Advancing Cognitive Science with Language Models}

\title[Article Title]{Addressing Longstanding Challenges in Cognitive Science with Language Models}

\author*[1,2]{\fnm{Dirk U.} \sur{Wulff}}\email{wulff@mpib-berlin.mpg.de}

\author*[2]{\fnm{Rui} \sur{Mata}}\email{rui.mata@unibas.ch}

\affil*[1]{\orgdiv{Center for Adaptive Rationality}, \orgname{Max Planck Institute for Human Development}, \orgaddress{\street{Lentzeallee 94}, \city{Berlin}, \postcode{14195}, \state{Berlin}, \country{Germany}}}

\affil[2]{\orgdiv{Faculty of Psychology}, \orgname{University of Basel}, \orgaddress{\street{Missionsstrasse 60/62}, \city{Basel}, \postcode{4055}, \state{Basel}, \country{Switzerland}}}

\maketitle

\section*{Abstract}

Cognitive science faces ongoing challenges in research integration, formalization, conceptual clarity, and other areas, in part due to its multifaceted and interdisciplinary nature. Recent advances in artificial intelligence, particularly the development of language models, offer tools that may help to address these longstanding issues. Specifically, they can help map fragmented literatures, formalize verbal theories, identify overlap among constructs and measures, generate predictions across tasks, and extract cultural or ecological structure from naturalistic data. However, these opportunities come with risks, including oversimplification, opacity, deskilling, and bias. Taken together, we conclude that language models could serve as tools for a more integrative and cumulative cognitive science when used judiciously to complement, rather than replace, human agency.

\vspace{1em}
\noindent\textbf{Keywords:} Large language models, cognitive science, conceptual clarity, formalization, measurement


\section{How Language Models Can Advance Cognitive Science}

Since its inception, cognitive science has aimed to unify insights from philosophy, psychology, neuroscience, computer science, and other disciplines to understand the mind \cite{gardner_minds_1985}. Yet some scholars have identified challenges that can hinder this interdisciplinary vision. Critics have pointed to the fragmentation of the field into disciplinary and methodological silos \cite{nunez2019happened}, an overreliance on vague or verbal theories \cite{van_rooij_formalizing_2020,guest2021computational}, the proliferation of redundant constructs and measures \cite{anvari_defragmenting_2025}, a lack of integrative modeling frameworks capable of generalization across tasks \cite{newell_you_1973,anderson_integrated_2004, yarkoni_generalizability_2022}, and limited attention to contextual and individual variation \cite{barrett_towards_2020, Henrich2010, wig_participant_2024}. In this paper, we examine how emerging applications of language models in cognitive science may help address these longstanding challenges. Although scholars may differ in how pressing they consider each of these concerns \cite{sanbonmatsu_there_2025}, recent developments in the field invite renewed reflection on how they might be tackled.

\begin{figure*}[t]
 \centering
 \includegraphics[width=1\linewidth]{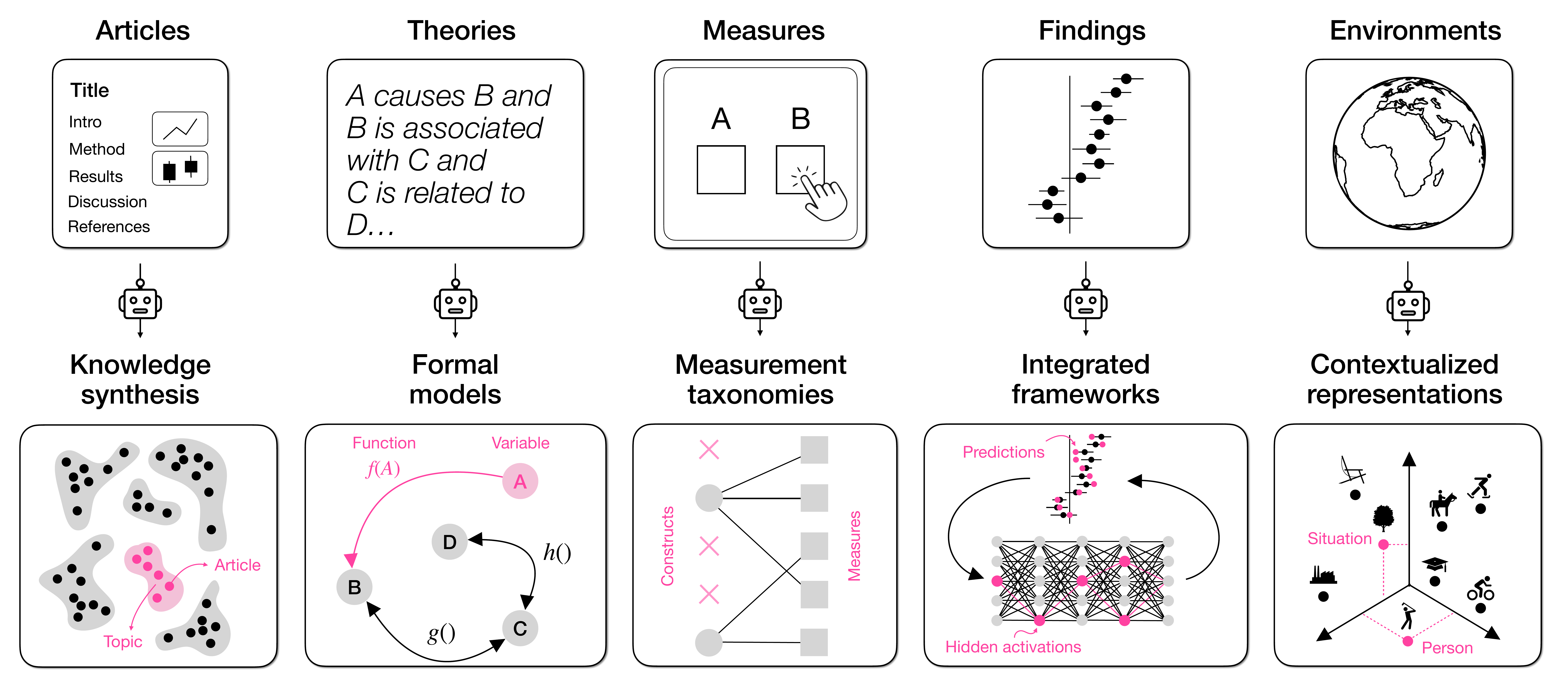}
 \caption{Leveraging language models to address longstanding challenges in cognitive science. From left to right, the five columns correspond to example research inputs or foci (Articles, Theories, Measures, Findings, Environments) and how these can be processed using language models to produce useful outputs. \textbf{Knowledge synthesis}: Language models help embed and index research articles, producing semantic maps that synthesize topics and reveal cross-field connections, thereby identifying research silos. \textbf{Formal models}: Language models assist in translating verbal theories into formal or executable models for clearer assumptions and testable predictions. \textbf{Measurement taxonomies}: embeddings from language models help to align measures with constructs, detect redundancy, and support principled relabeling. \textbf{Integrated frameworks}: Language models support generalizable prediction across tasks to provide accounts of empirical findings. \textbf{Contextualized representations}: Language models help capture ecological, cultural, situational, and individual variation from real-world contexts to improve context-sensitive representations.}
 \label{fig:approach}
\end{figure*}

We should note that the challenges we discuss may not be fully unique to cognitive science, but may reflect common difficulties associated with interdisciplinary fields, particularly those concerned with human behavior. Their relevance to cognitive science differs in kind and degree. Disciplinary fragmentation poses a foundational tension for cognitive science because the field was explicitly founded to integrate insights across disciplines and levels of analysis. Insufficient formalization and conceptual and measurement confusion are especially relevant because many central targets of cognitive science are latent constructs inferred from diverse behavioral, neural, and computational indicators, making them vulnerable to divergent operationalizations and construct proliferation. Limited generalizability and neglect of ecological context and variation are broader challenges shared with many sciences of human behavior, but they constrain cognitive science’s ability to explain minds and behavior across tasks, populations, and environments. Our aim is not to claim that these challenges are unique to cognitive science or equally distinctive within it, but to show why their combination is consequential for the field’s integrative ambitions.

In the following sections, we outline how language models (Box 1) can contribute to tackling these challenges (Figure 1). In some cases, language models are used primarily as tools, helping to reduce research silos via knowledge synthesis or contributing to theory formalization and measurement refinement. In other cases, they are used as cognitive models, assisting the provision of generative and generalizable predictions about human behavior and thought. They may also be used as models of the broader environment, helping to characterize cultural and ecological regularities and variation through contextualized representations. We focus on emerging examples from the cognitive sciences, while noting where the field can learn from developments elsewhere, given that some of these challenges are not unique to cognitive science. Throughout, we emphasize that language models should be viewed as supporting instruments rather than comprehensive solutions, and we conclude with a critical reflection on potential pitfalls.

\FloatBarrier
\begin{mdframed}

\subsection*{Box 1: Types of Language Models}\label{box:2}

Language models are computational systems trained on text to learn statistical regularities in language. The term often refers to models that predict the next linguistic unit given prior context. In a broader sense, which we adopt here, it includes models that encode linguistic input into structured numerical representations, such as word or sentence embeddings. Large language models (LLMs) are a contemporary subclass of language models characterized primarily by scale: they are trained on vast corpora, typically using transformer architectures \cite{vaswani2017attention}, and can perform a wide range of tasks without task-specific training. Thus, all LLMs are language models, but not all language models are large. Architecturally, language models differ in how they process text. Encoder models map input text to dense vector representations and are commonly trained using objectives such as masked token prediction or similarity learning \cite{devlin2019bert, reimers2019sentence}. Decoder models are optimized to generate text sequentially by predicting the next token \cite{brown2020language}. Encoder–decoder models combine both strategies. These architectural differences shape how models are used in scientific workflows, including clustering, prediction, generation, and simulation. Beyond architecture, models vary in training and adaptation. Base models capture general language structure through large-scale pretraining, whereas fine-tuned models are adapted to specific tasks, often incorporating human feedback \cite{ouyang2022training}. More recent reasoning models apply reinforcement learning over reasoning traces or verifiable outcomes to improve multi-step problem solving \cite{guo2025deepseek}. Finally, multimodal models extend language modeling to integrate text with images, audio, or video \cite{team2023gemini}. The development of modern language models builds on a long research tradition in the cognitive sciences, from probabilistic analysis of word sequences \cite{shannon1951prediction} to the distributional hypothesis \cite{firth1957synopsis, harris1954distributional} and its implementation in neural networks within the connectionist tradition \cite{rumelhart86, mcgrath_2024}.

For scientific applications, openness is a further critical dimension. Fully open models allow inspection of architecture, weights, and training sources \cite{palmer2024using}. Open-weight models provide downloadable parameters but, like their closed cousins, often have limited transparency regarding training \cite{liesenfeld2024rethinking} and are commonly distributed via repositories such as Hugging Face \cite{hussain2024tutorial}. Open systems promote transparency, reproducibility, and adaptation, and have already supported work on semantic measurement taxonomies \cite{wulff2025semantic} and predictive behavioral modeling \cite{binz_foundation_2025}. Nevertheless, much behavioral research relies on closed models \cite{wulff2024behavioral}. Requiring justification for model choice and investing in open infrastructures may help align model selection with scientific priorities rather than convenience \cite{palmer2024using, hussain2024tutorial}.

\end{mdframed}

\section{Knowledge Synthesis}

Productive interaction among disciplinary subcultures and approaches is a general challenge for the sciences \cite{cronbach_1957, snow1959two, NAP18722}, but it has long had special significance for cognitive science \cite{gardner_minds_1985}. Yet substantive integration across the cognitive sciences remains limited \cite{jacobs2014defense, nunez2019happened}. For example, research on decision making spans psychology, neuroscience, economics, and computational modeling, but these traditions often use different constructs, tasks, and explanatory vocabularies, making integration difficult \cite{frey2017risk,hertwig2019three}.

There are several ways in which language models can support integration through \textbf{knowledge synthesis}. First, language models can help efficiently map research areas within the cognitive sciences, giving researchers a rapid view of how their work relates to existing literature, including strands of work that are not immediately apparent because they arise in other disciplines or use different terminology \cite{cantone_estimation_2025}. Whereas earlier mapping tools primarily relied on citation networks or surface-level bibliometric patterns \cite{green_digital_2016}, language models can help surface deeper conceptual relations, particularly when human coding is imperfect \cite{tornberg_best_2024} or not feasible on a large scale \cite{nishikawa2024llm_citation}. Emerging applications suggest that such tools can map specific research areas in cognitive science, such as reinforcement learning, to detect silos or cross-domain linkages to uncover opportunities for collaboration \cite{taher2023embedding, thoma2025mapping}.

Second, language models can be used to test the coherence and cumulative strength of these research landscapes through prediction. Pipelines integrating language models can generate out-of-sample predictions, identify candidate predictors, and benchmark competing theories \cite{savcisens_using_2023, luo2024large}. For example, language models have been shown to predict study outcomes in different neuroscience domains more accurately than human experts \cite{luo2024large}. Such workflows may help identify gaps in current knowledge and provide reference points for cumulative progress, especially when combined with research maps that reveal where linked domains or subfields do or do not yield generalizable insights \cite{musslick2025automating}.

Third, language models may accelerate systematic reviews and meta-analyses in both cognitive science and more broadly by supporting document triage and structured evidence extraction \cite{cao_automation_nodate, babaei_giglou_llms4synthesis_2024}. By making findings across domains visible, such tools can help to bridge fragmented literatures and facilitate cross-disciplinary integration in many areas of science.

Automated mapping, prediction, and synthesis can reveal connections across fields and assess the generalizability of existing evidence, but they do not guarantee integration. Progress will therefore depend on researchers who translate automated outputs into cumulative frameworks.

\section{Formal Models}

Although \textbf{formal models} are important across the sciences \cite{frigg_models_2018}, the vagueness and imprecision of many theories of the mind have long made formalization a concern in cognitive science \cite{meehl_1990}. Numerous scholars have called for greater formalization to enhance theoretical clarity, rigor, and testability \cite{oberauer_addressing_2019, smaldino_how_2020, van_rooij_formalizing_2020, guest2021computational}. Several barriers to this vision remain. Many researchers lack sufficient training in formal modeling, and the volume and velocity of contemporary scientific output make large-scale formalization that keeps pace with new findings difficult without automated support \cite{michie_human_2017}.

In cognitive science and adjacent areas, language models are already being used to address these pragmatic barriers. One key application is translating verbal theories into symbolic or executable code, especially in domains that have traditionally lacked formalization \cite{Read_Monroe_2023}. For example, researchers have proposed using language models to help identify causal relationships between variables when constructing causal models. In psychopathology, a language model can be prompted to suggest directional relationships between pairs of symptoms, which are then assembled into a broader causal structure \cite{waaijers_theoraizer_2024}. These tools could reduce human effort and time when applied at scale \cite{crielaard_refining_2024}. Of course, these efforts require validation, and human experts will ultimately need to determine the validity and usefulness of such formal theories. 

Language models can also be used to generate computational models in domains that already employ mathematical or other formal (e.g., algorithmic) approaches. In this area, an open question is whether automated methods create novel models or primarily rediscover existing ones \cite{peterson_using_2021}. However, recent findings suggest that language models can be generative. In one example, researchers developed a pipeline that prompts language models to propose computational models from task descriptions and participant data, and then iteratively refines the models using performance feedback on held-out data \cite{rmus_generating_2025}. The results show that this approach can produce well-performing models across domains (i.e., decision making, learning, planning, working memory). This approach also yielded models that differed from existing ones, indicating novelty, and in many cases matched or surpassed established computational models in predictive performance in the tested domains, which suggests that computational models generated using language models can exceed current benchmarks in a number of areas. A complementary demonstration comes from another group of researchers, who formalized dozens of reasons as explicit choice functions and used a language model to map participants’ natural language explanations onto these formal rules \cite{fulawka2025reasons}. Their approach captured systematic heterogeneity in the application of decision reasons. This variation accounted for people's choices better than classical models, such as prospect theory, illustrating how language models can help improve models of human decision-making.

Although these examples demonstrate that language models can help to formalize theories and generate new computational models, their use also raises epistemic risks. Lowering the barriers to formalization could lead to the proliferation of superficially formal but theoretically weak models, underscoring the continued importance of expert evaluation. Questions remain about the interpretability of increasingly complex models, the extent to which automation might deskill researchers, and the need to ensure accountability for model outputs. These broader issues are discussed in the section on potential pitfalls. 

\section{Measurement Taxonomies}

\begin{figure*}[t]
 \centering
 \includegraphics[width=1\linewidth]{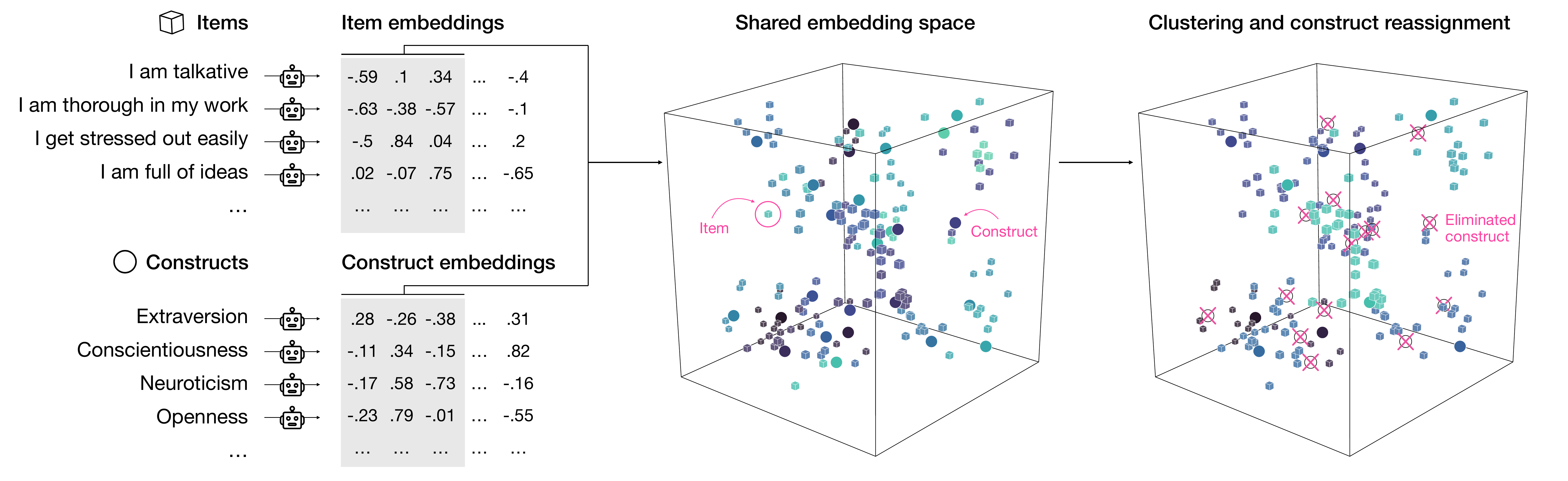}
 \caption{Embedding-based mapping and relabeling of psychological measures.
The figure illustrates how embeddings can be used to place questionnaire items and construct labels in a shared semantic space, reveal conceptual overlap, and reduce redundancy. Left: Individual items and construct labels are encoded as high-dimensional vectors derived from language models. Center: These vectors are projected into a common embedding space in which proximity reflects semantic similarity; items and constructs that cluster together likely capture overlapping meaning. (The cube depicts a 3D schematic; in practice, embeddings have many more dimensions.) Right: Clustering within this space supports systematic relabeling or consolidation: Constructs with highly similar item profiles can be reassigned or eliminated, yielding a more parsimonious taxonomy of measures and associated constructs.}
 \label{fig:embedding}
\end{figure*}

The need for conceptual engineering, that is, the systematic assessment and refinement of scientific concepts over time, is not unique to cognitive science \citep{chalmers_what_2020}. However, the problem appears particularly pronounced in cognitive science, where limited communication between disciplines and subfields has contributed to the proliferation of theories, constructs, and measures \cite{anvari_defragmenting_2025}. Analysis of the American Psychological Association’s PsycTests database identified more than 38,000 distinct constructs and an even greater number of unique measures, with a large portion only having been used once or twice \cite{elson_psychological_2023}. This level of fragmentation impedes cumulative progress, leaving researchers struggling to select appropriate measures, assess novelty, and build unified intervention frameworks. The resulting conceptual and measurement sprawl has prompted calls for deliberate consolidation efforts \citep{eronen2021theory}. Although structured \textbf{taxonomies} and \textbf{ontologies} have been proposed \cite{poldrack_brain_2016, sharp_use_2023}, these initiatives remain incomplete and themselves require integration as their numbers increase \cite{norris_scoping_2019}.

Language models offer promising support with the analysis of extensive textual corpora in order to identify redundant or overlapping constructs, cluster semantically related terms, and propose more coherent taxonomies or ontologies. There is a growing tradition of efforts to improve concepts and measurement using automated language-model–based methods \cite{larsen2016tool, rosenbusch2020semantic}, and recent advances show \textbf{semantic embeddings} from language models can be used to capture key links between measures and constructs \cite{hommel2024language, wulff2025semantic}. For example, drawing on embeddings of items, scales, and construct labels learned from a large corpus of personality instruments, researchers modeled the semantic landscape linking thousands of questionnaire items to hundreds of higher-level constructs (Figure ~\ref{fig:embedding}) \cite{wulff2025semantic}. These representations reproduced empirical item–scale relationships well and were used to flag problematic matches between scales and constructs to tackle so-called \textbf{jingle–jangle fallacies}. The authors also outlined procedures to prune and reorganize taxonomies by reallocating labels to scales to reduce conceptual and measurement overlap. In one demonstration, they sketched a condensed personality framework that reduced the hypothesized construct set by roughly 75\%, illustrating how semantic embeddings can support more economical and internally consistent measurement taxonomies.

The work described above has largely focused on text-based measures, such as personality items from self-reports, which align well with the linguistic information captured by language models. However, recent work suggests that language models may also be used to predict behavioral outcomes for task-based measures \citep{binz_foundation_2025}, indicating that such models could, at least in principle, be used to integrate data from different measurement approaches to help build a more comprehensive map of psychological measurement.

Language models can also assist in designing, populating, and integrating larger knowledge structures, such as ontologies. This type of emerging application, often termed ontology learning, aims to automate the time-consuming, typically expert-driven process of knowledge structuring. For instance, researchers are actively testing language models on core tasks such as discovering taxonomic hierarchies and semi-automatically constructing new ontologies from scholarly texts \cite{sadruddin2025llms4schemadiscovery, babaei2023llms4ol}. Furthermore, language models are being applied to ontology matching; that is, the task of identifying correspondences between different, heterogeneous knowledge structures \cite{giglou2024llms4om}. Such tools are vital for consolidating the fragmented conceptual landscape by helping to formally integrate the field's redundant constructs and measures as they develop into comprehensive ontologies \cite{norris_scoping_2019}.

Although language models offer powerful means to identify redundancies and promote more coherent taxonomies, automated consolidation of constructs and measures also poses some challenges. Questions remain about how to balance conceptual clarity with theoretical diversity, avoid reinforcing dominant frameworks, and ensure transparency in decisions about which constructs are retained or redefined \cite{wulff_mata_2025}. Although some skepticism may be warranted, recent progress in large-scale changes to scientific practice in the cognitive sciences \cite{chambers_past_2021}, together with growing engagement with broader questions of conceptual clarity \cite{anvari_defragmenting_2025}, suggests cautious optimism.

\section{Integrated Frameworks}

\begin{figure*}[t]
 \centering
 \includegraphics[width=1\linewidth]{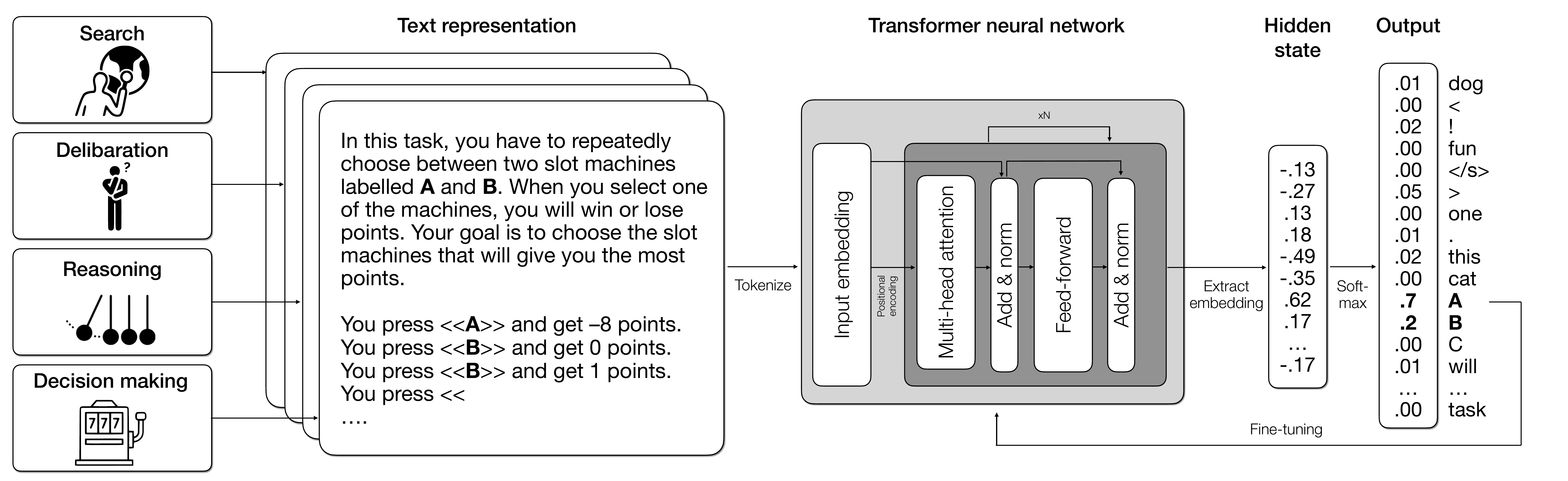}
 \caption{The figure illustrates the approach underlying Centaur \cite{binz_foundation_2025}, a foundation model of human cognition, trained to predict behavior across diverse experimental tasks. Left: Task instructions, stimuli, and participants’ trial histories from different cognitive aspects (search, deliberation, reasoning, decision making) are first translated into text and then tokenized to serve as model input. The corresponding input embeddings pass through a transformer neural network architecture (embedding layer, multi-head attention, and feed-forward blocks) to produce context-sensitive hidden representations of the task state. The model then outputs a probability distribution over possible outputs, including those representing task actions (for example, which option a participant will choose next) using a softmax layer. This approach has been used to capture behavioral regularities across multiple tasks (e.g., digit span, two-armed bandit), and it has been shown to generalize to new task structures and domains, implying that it may represent a unified framework for predicting and interpreting human behavior.}
 \label{fig:centaur}
\end{figure*}

The need for models that generalize beyond narrow training or testing contexts is not unique to cognitive science \cite{marcellesi_external_2015}, but the critique of one model per phenomenon has long shaped debate in the cognitive sciences \cite{newell_you_1973}. It points to the field’s historical tendency to develop narrowly scoped models that focus on specific experimental paradigms or domains that rarely generalize beyond their immediate application. In response, researchers have called for integrated cognitive architectures that can explain and predict behavior across multiple tasks and domains \cite{anderson_integrated_2004, LAIRD19871}. However, these architectures have also been criticized for being handcrafted, requiring extensive task-specific parameterization, and lacking systematic empirical validation \cite{sun_introduction_2001, byrne_unified_2012}. A related concern is that existing approaches have tended to prioritize post hoc explanation over predictive accuracy, thereby limiting their real-world applicability \cite{yarkoni_choosing_2017}. Addressing these limitations requires models equipped to provide scalable and generalizable prediction across diverse cognitive phenomena.

Neural network architectures related to those underlying language models are already part of a comparatively mature and empirically rich literature on cognitive modeling. Further, a growing body of work uses language models to investigate various aspects of human cognition \cite{frank_cognitive_2025}. At the same time, much of the existing literature has made substantial progress within particular domains, especially language processing \cite{de_varda_cost_2025, lampinen_language_2024} and brain-model alignment \cite{tuckute_language_2024, schrimpf_neural_2021}. However, newer generations of language modules promise to further advance \textbf{integrated modeling frameworks} by providing models that generalize across tasks, domains, and experimental settings. As multitask learners trained on broad and diverse data, language models can be viewed as generalist architectures capable of performing a wide variety of cognitive and behavioral tasks, potentially without task-specific redesign.

Recent research work exemplifies the breadth of this approach with Centaur, a \textbf{foundation model} trained on an extensive collection of behavioral datasets spanning decision making, learning, memory, and cognitive control tasks (Figure~\ref{fig:centaur}) \cite{binz_foundation_2025}. Centaur encodes task structures, stimuli, and behavioral responses into a shared latent representation, allowing it to predict human behavior across a wide range of experimental conditions. The model has been used to explain substantial variance in human responses and generalizes to unseen tasks and domains, outperforming traditional task- or domain-specific models. Moreover, Centaur’s internal representations appear to align with neural activity patterns, suggesting that large-scale multitask behavioral modeling could yield mechanistically informative representations of cognition. Continued advances toward multimodal and interpretable architectures may allow future foundation models to achieve more comprehensive and predictive accounts of cognition.

Although these developments illustrate the promise of language models for building unified models of cognition, several challenges remain. Current systems, specifically Centaur, can be brittle and sensitive to small variations in task input \cite{kieval_captured_2025, xie2025centaur, schroder2025large}, and their internal representations often remain opaque, complicating interpretation and theoretical insight \cite{frank_cognitive_2025}. Others have more fundamentally questioned the potential of any currently available (artificial) model to advance understanding of human cognition \cite{van2024reclaiming}. These issues, along with broader methodological risks, are further discussed in the section on potential pitfalls. 

\section{Contextualized Representations}

Across the human and behavioral sciences, concerns about external validity center on whether findings from one sample or context generalize to broader or different populations \cite{findley_external_2021}. This issue is particularly central in cognitive science because the field aims to explain minds and behavior across individuals, cultures, and environments, yet much psychological theory and experimentation have been developed in controlled laboratory environments that prioritize internal validity over scope and \textbf{ecological validity} \cite{barrett_towards_2020}. Cross-cultural research has shown that many psychological constructs and effects fail to generalize beyond the \textbf{WEIRD populations} (Western, Educated, Industrialized, Rich, and Democratic)  from which most data are drawn \cite{wig_participant_2024, Henrich2010}. This lack of representativeness limits their applicability to real-world applications. Addressing these challenges requires methods that can incorporate naturalistic, situated, and demographically diverse data into theory, modeling, and prediction.

We see two main ways in which language models can support contextualized representations: by enabling large-scale extraction of contextual and person-related features from naturalistic and situated data, and by supporting exploratory simulations of contextual and demographic variation \textit{in silico}.

Trained on vast, heterogeneous corpora drawn from real-world language use, language models could help capture patterns that potentially reflect cultural norms, social practices, and situational variability across contexts and even historical time \cite{varnum_large_2024}. In a recent example, researchers used a pipeline including language models to process over 100,000 real-life choice dilemmas from online forums, successfully extracting the underlying decision attributes and trade-offs from unstructured text \cite{bhatia2025computational}. This type of approach allows quantifying how personal and social considerations vary across different contexts, thereby enhancing ecological coverage and creating contextualized representations of various real-life choices \cite{yudkin_large-scale_2023}. Similar opportunities will likely arise for analyzing other data types as language models increasingly gain multimodal capacities, integrating text with images, audio, and other modalities within a single representational space. 

Second, through persona-based \textbf{steering}, language models can be used to simulate responses from diverse demographic or social groups \cite{dillion_can_2023, lutz2025prompt, haller2024opiniongpt}, albeit with important limitations and risks \cite{wang_large_2025, crockett_ai_2025}. Such simulations may provide a provisional means of exploring how existing theories or interventions might generalize across contexts when direct data collection is limited or infeasible, for example, when incorporated into agent-based simulations to model heterogeneous populations \cite{gao_large_2024}. Importantly, these simulations should not be viewed as substitutes for empirical engagement with real populations, but as exploratory tools that help make assumptions visible and identify where further data collection is most needed.

Despite these possibilities, contextualized modeling with language models currently faces significant limitations. Models trained on real-world data are only as representative as their underlying corpora, which often overrepresent dominant linguistic and cultural groups while underrepresenting marginalized or low-resource populations \cite{sen_2025, havaldar_building_2024}. Also, research shows that language model outputs can require fine-tuning and vary substantially depending on model version and prompting strategy \cite{haller2023opiniongpt,suh2025language}, especially in applications such as demographic steering through the use of personas \cite{cummins_2025, tosato2025persistent}. As a result, realizing the potential of contextualized representations will require deliberate efforts to diversify training data and rigorously validate model outputs across cultural and demographic contexts.

\begin{table}[h]
\footnotesize
\centering
\begin{tabular}{p{2.0cm} p{3.2cm} p{3.8cm} p{3.2cm}}
\textbf{Challenge} & \textbf{Description} & \textbf{LM-supported Solutions} & \textbf{Human Oversight} \\
\hline
Disciplinary silos & Different disciplines and subfields do not coalesce in their efforts, leading to conceptual and methodological silos \cite{cronbach_1957, nunez2019happened} & Develop cross-disciplinary mapping tools: Language models can help to construct research maps that reveal latent conceptual and methodological overlaps across fields \cite{thoma2025mapping} and assess their predictive utility \cite{luo2024large} & Interpreting mappings, evaluating their theoretical significance, and deciding which integrations are scientifically warranted \\
Insufficient formalization & Overreliance on vague and verbal theorizing and limited training in formal modeling lead to a lack of formal theories and clear predictions \cite{oberauer_addressing_2019, smaldino_how_2020}  & Promote testable, formal theories: Language models can assist in translating verbal theories into symbolic or executable code \cite{rmus_generating_2025, waaijers_theoraizer_2024} & Choosing modeling frameworks, evaluating assumptions, validating predictions, and assessing theoretical adequacy \\
Conceptual and measurement confusion & Unchecked proliferation of constructs and measures, leading to redundancy and ambiguity (e.g., jingle--jangle fallacies) \cite{anvari_defragmenting_2025, poldrack_brain_2016} & Consolidate psychological constructs and create measurement taxonomies: Language models can analyze corpora of texts and measures to identify overlapping constructs, cluster semantically related ones, and propose more coherent taxonomies of measures \cite{wulff2025semantic}. & Defining constructs, documenting consolidation criteria, and making normative decisions about validity, diversity, and theoretical scope \\
Lack of generalizability & Models are often narrow and task-specific, with poor generalization across tasks \cite{newell_you_1973, yarkoni_generalizability_2022} & Develop multitask models and unified cognitive architectures: Language models, as multitask learners, provide a platform for assessing generalist capabilities that can be probed to advance knowledge of computational principles and make behavioral predictions \cite{binz_foundation_2025}  & Framing hypotheses, interpreting successes and failures of generalization, drawing theoretical conclusions \\
Neglect of ecological context and variation & Theories often omit ecological, cultural, and individual variation, reducing validity \cite{barrett_towards_2020, wig_participant_2024} & Integrate ecological and contextual aspects: Language models can help process and extract meaningful patterns from real-world, naturalistic datasets (e.g., social media), helping researchers to account for ecological, cultural, historical, or individual differences \cite{varnum_large_2024, bhatia2025computational} and enable in-silico testing of interventions across diverse populations \cite{gao_large_2024} & Ethical judgment, contextual grounding, validating subgroup performance, and deciding what variability is theoretically meaningful \\
\hline
\end{tabular}
\caption{Historical challenges, language model (LM)-supported solutions, and primary human oversight roles}
\label{tab:conceptual_clarity_table}
\end{table}

\section{Avoiding Potential Pitfalls}

Although language models may help address persistent challenges in cognitive science, important risks remain. Their benefits will depend not only on technical capabilities but also on how they are embedded within scientific norms, incentives, and safeguards. Below, we outline major risks and practical mitigation strategies before considering possible futures for their use.

\subsection{Major risks and mitigation strategies}

We focus on four risks—oversimplification, opacity, deskilling, and bias—that are especially relevant to the uses discussed above. Table 1 summarizes the challenges discussed, possible language-model-supported solutions, and the human oversight roles that remain central to realizing these benefits.

\textbf{Oversimplification}. The drive for coherence and unification, although essential to cumulative science, can slip into reductionism. Cognitive phenomena are complex and context-dependent, and not all ambiguity reflects theoretical failure \cite{sanbonmatsu_there_2025}. When automation prioritizes simplicity and uniformity, constructs (e.g., intelligence, psychopathology) risk being flattened into one-size-fits-all templates, and theoretical pluralism can be replaced by a false consensus \cite{hochstein_categorizing_2016, danziger_psychology_2013}. To avoid oversimplification, synthesis and integration efforts should remain transparent and revisable. This may involve explicitly documenting criteria for construct merging, maintaining versioned records of alternative conceptualizations, involving experts from multiple subfields in review processes, and routinely evaluating whether harmonized constructs preserve meaningful variation across contexts and populations \cite{wulff_mata_2025}. Encouragingly, examples of best practices are developing \cite{leising_tentative_2024}, with the challenge being how to incorporate input from language models into consensus building  while preserving the pluralism and integrative spirit that have long defined cognitive science \cite{gentner_cognitive_2019}. In this regard, cognitive science may be able to learn from other fields, such as public health, that are also developing approaches for integrating multiple perspectives and models into forms of consensus building \cite{shea_harnessing_2020}.

\textbf{Opacity}. Language models can already achieve strong predictive performance across behavioral tasks, but good predictions alone do not guarantee explanatory value \cite{narayanan2025overreliance, frank_cognitive_2025, van2024reclaiming}. This is especially problematic when language models are treated not merely as tools for synthesis or prediction, but as candidate models of cognition. Models may rely on statistical shortcuts or biases in training data that are not obvious from their outputs \cite{van2025combining, buijsman_epistemic_2024}. In addition, apparent predictive success may be inflated when models have been exposed to related papers, tasks, or benchmarks during training, making independent validation and transparent model documentation essential \cite{balloccu2024leak, barrie2025emergent, ashery2025reply}. This is one reason why openness and model choice matter for scientific applications (Box 1). Avoiding misplaced explanatory claims requires treating interpretability as a methodological requirement alongside predictive performance, especially for integrated modeling frameworks \cite{mcgrath_2024}. Cognitive science can contribute here by adapting experimental tools such as functional localization and lesion methods to probe the internal organization of language models and move toward causal accounts of model behavior \cite{demircan2024sparse, hussain2024probing, zhu_using_2025, alkhamissi2025llm}.

\textbf{Deskilling}. Closed or highly automated infrastructures can erode human competencies \cite{musslick2025automating}. When researchers rely on opaque systems to define constructs or generate models, they risk becoming operators of tools they cannot interrogate \cite{van2025combining}. Over time, this could lead to increased dependence on automated methods and loss of capacity \cite{ferdman_ai_2025}. Yet, not all forms of automation are detrimental. Delegating routine tasks could free resources for theory building and critical evaluation. Avoiding harmful deskilling may require deliberate capacity-building efforts, such as “manual-first” training that keeps conceptual reasoning and model construction at the center of undergraduate and graduate education, as well as cultivating new strategies that ensure evaluation of intermediate representations and support human-in-the-loop workflows. Ultimately, the future of the field will depend on whether expertise evolves alongside the technologies it relies on.

\textbf{Bias}. Both cognitive science datasets and language model training corpora are heavily skewed toward Western populations and underrepresent marginalized groups \cite{wig_participant_2024, barrett_towards_2020, santurkar2023whose, sen_2025}. When such data are used for training, analysis, or simulation, there is a risk of overlooking, reproducing, or even amplifying existing cultural and linguistic biases \cite{crockett_ai_2025}. Mitigating these risks requires deliberate action, including diversifying training and evaluation data, documenting data provenance, conducting systematic bias audits, and validating model outputs across demographic subgroups and ecological contexts.

\subsection{Possible Futures}

Taken together, these considerations suggest that the long-term impact of language models will depend less on their technical capabilities than on the norms and institutional practices that shape their use. To clarify what is at stake, we outline two contrasting trajectories for integrating language models into cognitive science, inspired by scenario-based approaches from AI governance \cite{cave_hopes_2019}.

In a dystopian future, the infrastructures built to organize knowledge end up constraining it. Tools that initially helped scientists navigate a complex literature now dictate which questions are worth asking and which findings are deemed relevant. Automated synthesis amplifies what is already well represented while obscuring novelty and dissent. Theories are formalized automatically, but their assumptions become opaque, and models predict well but have limited explanatory power. Measurement systems are streamlined for computational convenience, collapsing the diversity of constructs into standardized templates, narrowing conceptual diversity. Unified frameworks achieve generality and are prized for predictive validity, but their internal workings grow too complex to interpret and may be too locked within proprietary architectures to allow investigation. In this trajectory, research accelerates, but at the cost of interpretability, pluralism, and sensitivity to contextual variation.

In a more desirable future, the same tools foster a more reflective and integrative cognitive science. Automated synthesis serves as a compass, not a fixed path, helping researchers to trace conceptual connections and uncover neglected questions across subfields and disciplinary boundaries. Formalization is accessible and collaborative, with language models supporting the translation of ideas into precise, testable forms while keeping assumptions transparent and revisable by human researchers. Redundancy in measures and constructs is reduced, but diversity of perspectives is preserved. Predictive frameworks are both generalizable and interpretable, advancing understanding by linking performance to mechanism rather than replacing explanation. Representations of cognition expand to include the variability of real-world contexts, acknowledging that minds differ across environments, cultures, and history. Here, automation enhances human reasoning, and conceptual clarity emerges from openness, pluralism, and sustained dialogue between researchers.

Which future prevails will depend on whether language models are simply used to produce results faster or as supporting tools for deliberate, critical, and responsible scientific practice.

\section{Concluding Remarks}

We have argued that language models can be used to help address longstanding challenges in cognitive science. However, for a successful outcome, they must be embedded in transparent and pluralistic research practices to expand the scope and scale of inquiry while supporting integrative forms of explanation. All in all, their value lies not in full automation, but in how they are used to strengthen a cumulative, interpretable, and context-sensitive science of the mind.

\begin{mdframed}
\subsection*{Box 2: Outstanding Questions}\label{box:3}

\begin{enumerate}
    \item How can we best validate the accuracy and completeness of knowledge synthesis (e.g., research maps) automatically extracted by language models from scientific literature?
    \item When language models assist in formalization by generating novel computational models, what methods can ensure that these models are not just predictive ``black boxes" but also provide interpretable, mechanistic insights that advance human theoretical understanding?
    \item What are the fundamental limitations of using models trained predominantly on text to understand, simulate, or automate research related to cognitive processes deeply grounded in embodiment, perception, and action?
    \item What validation frameworks and ethical guidelines are necessary to ensure the reliability and responsible use of language models in automating different research stages such as hypothesis generation or outcome prediction?
    \item How do specific architectural choices, training regimes, and data compositions within open-source language models impact their suitability and potential biases when applied to different cognitive science problems?
    \item Can language model-assisted generation of cognitive models lead to truly novel theoretical frameworks, or does it primarily accelerate the exploration within existing paradigms?
    \item What infrastructural, educational, and collaborative frameworks are needed to ensure that language model-based tools are effectively implemented, maintained, and equitably accessible across the cognitive science community?
    \item What new scientific and educational practices are necessary to mitigate the risk of deskilling? How do we train researchers to use language models as complements that enhance critical and theoretical expertise?
\end{enumerate}
\end{mdframed}

\newpage

\section{Highlights}

\begin{itemize}
    \item Language models offer new tools to address longstanding challenges in cognitive science.
    \item We outline key use cases, from helping reduce research silos, supporting formalization, consolidating measurement taxonomies, enabling integrative modeling frameworks, and capturing contextual variation.
    \item The promise of language models is contingent on their responsible use; they should complement human expertise and be deployed with safeguards against risks such as oversimplification, opacity, deskilling, and bias.
\end{itemize}

\newpage

\section{Glossary}

\begin{itemize}
    \item \textbf{Conceptual Engineering}: The practice of assessing and improving our concepts to better serve our scientific or practical goals.
    \item \textbf{Ecological Validity}: The extent to which the findings of a research study may be generalized to real-life settings.
    \item \textbf{Fine-tuned}: The process of taking a pretrained foundation model and further training it on a smaller, domain-specific dataset. This adapts the model to perform specialized tasks or to adopt a particular style or knowledge base.
    \item \textbf{Formal Models}: Theories expressed in a precise mathematical or computational language to eliminate ambiguity and allow for direct simulation and testing.
    \item \textbf{Foundation Model}: A large-scale AI model trained on a massive amount of broad data that can be adapted to a wide range of downstream tasks.
    \item \textbf{Integrated Modeling Frameworks (Cognitive Architectures)}: Broad, unified theories of cognition that aim to explain and predict behavior across a wide range of tasks and domains, rather than focusing on a single phenomenon.
    \item \textbf{Jingle--Jangle Fallacies}: The ``jingle" fallacy is the error of assuming two different things are the same because they have the same name, whereas the ``jangle" fallacy is the error of assuming two identical things are different because they have different names.
    \item \textbf{Knowledge Synthesis}: The process of integrating findings from different studies, disciplines, or sources to create a more comprehensive understanding of a topic.
    \item \textbf{Language Models}: Computational models that learn statistical regularities in sequences of linguistic units (e.g., words or tokens). They are trained to predict, generate, or represent language, often by estimating the probability of sequences or by encoding text into numerical representations (embeddings). Language models may serve different functions, including text generation, semantic similarity, clustering, and prediction.
    \item \textbf{Ontologies}: Formal representations of knowledge as a set of concepts within a domain and the relationships that hold between them.
    \item \textbf{Semantic Embeddings}: Numerical representations of words, sentences, or documents in a high-dimensional space where proximity corresponds to similarity in meaning.
    \item \textbf{Steering}: The method of providing a specific instruction, question, or context as input to a language model to guide its output. The design of the prompt is crucial for controlling the model's behavior and the quality of its response.
    \item \textbf{Taxonomies}: Structured systems for classifying and organizing entities within a domain according to shared properties or conceptual relationships. In cognitive science, taxonomies often organize psychological constructs and measurement instruments (e.g., questionnaires, tasks) to clarify conceptual boundaries, reduce redundancy, and support cumulative research.
    \item \textbf{WEIRD Populations}: An acronym for research participants from Western, Educated, Industrialized, Rich, and Democratic societies, who are overrepresented in psychological research.

\end{itemize}


\subsection*{Acknowledgments}

We acknowledge funding from the German Research Foundation to Dirk U. Wulff (546419617) and through grant HE 2768/11-1, and from the Swiss National Science Foundation to Rui Mata (204700). We thank Samuel Aeschbach, Anna Thoma, Taisiia Tikhomirova, Valentin Kriegmair, and Loreen Tisdall for helpful comments and Laura Wiles for editing the manuscript. 

\subsection*{Declaration of interests}

The authors declare no competing interests. \\ 

\subsection*{Declaration of Generative AI and AI-assisted technologies in the writing process}

During the preparation of this work, the authors used ChatGPT, Claude, and Grammarly to improve the manuscript's readability and language. After using these tools, the authors reviewed and edited the content as needed and take full responsibility for the content of the article.

\backmatter

\newpage

\bibliography{sn-bibliography}

@article{findley_external_2021,
  author  = {Findley, Michael G. and Kikuta, Kyosuke and Denly, Michael},
  title   = {External validity},
  journal = {Annual Review of Political Science},
  year    = {2021},
  volume  = {24},
  pages   = {365--393},
  doi     = {10.1146/annurev-polisci-041719-102556}
}

@article{marcellesi_external_2015,
  author  = {Marcellesi, Alexandre},
  title   = {External validity: Is there still a problem?},
  journal = {Philosophy of Science},
  year    = {2015},
  volume  = {82},
  number  = {5},
  pages   = {1308--1317},
  doi     = {10.1086/683534}
}

@incollection{frigg_models_2018,
  author    = {Frigg, Roman and Hartmann, Stephan},
  title     = {Models in science},
  booktitle = {The Stanford Encyclopedia of Philosophy},
  editor    = {Zalta, Edward N.},
  edition   = {Summer 2018},
  publisher = {Metaphysics Research Lab, Stanford University},
  year      = {2018},
  url       = {https://plato.stanford.edu/archives/sum2018/entries/models-science/}
}

@BOOK{NAP18722,
  author    = {{National Research Council}},
  title     = {Convergence: Facilitating transdisciplinary integration of life sciences, physical sciences, engineering, and beyond},
  isbn      = "978-0-309-30151-0",
  doi       = "10.17226/18722",
  url       = "https://nap.nationalacademies.org/catalog/18722/convergence-facilitating-transdisciplinary-integration-of-life-sciences-physical-sciences-engineering",
  year      = 2014,
  publisher = "The National Academies Press",
  address   = "Washington, DC"
}

@article{shea_harnessing_2020,
	title = {Harnessing multiple models for outbreak management},
	volume = {368},
	copyright = {http://www.sciencemag.org/about/science-licenses-journal-article-reuse},
	issn = {0036-8075, 1095-9203},
	url = {https://www.science.org/doi/10.1126/science.abb9934},
	doi = {10.1126/science.abb9934},
	language = {en},
	number = {6491},
	urldate = {2026-02-19},
	journal = {Science},
	author = {Shea, Katriona and Runge, Michael C. and Pannell, David and Probert, William J. M. and Li, Shou-Li and Tildesley, Michael and Ferrari, Matthew},
	month = may,
	year = {2020},
	pages = {577--579},
}

@article{yudkin_large-scale_2023,
	title = {A large-scale investigation of everyday moral dilemmas},
	copyright = {https://creativecommons.org/licenses/by/4.0/legalcode},
	url = {https://doi.org/10.1093/pnasnexus/pgaf119},
	doi = {10.1093/pnasnexus/pgaf119},
	language = {en},
	urldate = {2026-02-13},
	journal = {PNAS Nexus},
    volume = {4},
    number = {5},
	author = {Yudkin, Daniel Alexander and Goodwin, Geoffrey and Reece, Andrew Garrett and Gray, Kurt and Bhatia, Sudeep},
	month = May,
	year = {2025},
}

@article{gao_large_2024,
	title = {Large language models empowered agent-based modeling and simulation: A survey and perspectives},
	volume = {11},
	issn = {2662-9992},
	shorttitle = {Large language models empowered agent-based modeling and simulation},
	url = {https://www.nature.com/articles/s41599-024-03611-3},
	doi = {10.1057/s41599-024-03611-3},
	language = {en},
	number = {1},
	urldate = {2026-02-14},
	journal = {Humanities and Social Sciences Communications},
	author = {Gao, Chen and Lan, Xiaochong and Li, Nian and Yuan, Yuan and Ding, Jingtao and Zhou, Zhilun and Xu, Fengli and Li, Yong},
	month = sep,
	year = {2024},
	pages = {1259},
}

@article{dillion_can_2023,
	title = {Can {AI} language models replace human participants?},
	volume = {27},
	issn = {13646613},
	url = {https://linkinghub.elsevier.com/retrieve/pii/S1364661323000980},
	doi = {10.1016/j.tics.2023.04.008},
	language = {en},
	number = {7},
	urldate = {2023-10-05},
	journal = {Trends in Cognitive Sciences},
	author = {Dillion, Danica and Tandon, Niket and Gu, Yuling and Gray, Kurt},
	month = jul,
	year = {2023},
	pages = {597--600},
}

@article{tornberg_best_2024,
	title = {Best practices for text annotation with large language models},
	volume = {18},
	copyright = {Creative Commons Attribution 4.0 International},
	url = {https://sociologica.unibo.it/article/view/19461},
	doi = {10.6092/ISSN.1971-8853/19461},
	language = {en},
	number = {2},
	urldate = {2024-11-06},
	journal = {Sociologica},
	publisher = {Sociologica},
	author = {Törnberg, Petter},
	month = oct,
	year = {2024},
	pages = {67--85},
}

@article{cao_automation_nodate,
	title = {Automation of systematic reviews with large language models},
	language = {en},
	author = {Cao, Christian and Arora, Rohit and Cento, Paul and Manta, Katherine and Farahani, Elina and Cecere, Matthew and Sang, Jason and Gong, Ling Xi and Kloosterman, Robert and Jiang, Scott and Saleh, Richard and Margalik, Denis and Lin, James and Jomy, Jane and Xie, Jerry and Chen, David and Gorla, Jaswanth and Lee, Sylvia and Zhang, Kelvin and Ware, Harriet and Whelan, Mairead and Teja, Bijan and Leung, Alexander A and Ghosn, Lina and Arora, Rahul K and Detsky, Allen S and Noetel, Michael and Emerson, David B and Boutron, Isabelle and Moher, David and Church, George and Bobrovitz, Niklas},
    url = {https://doi.org/10.1101/2025.06.13.25329541},
    doi  = {10.1101/2025.06.13.25329541},
    journal = {medRxiv},
    year = {2025},
}

@article{cantone_estimation_2025,
	title = {Estimation of disciplinary similarity with large language models},
	volume = {130},
	issn = {0138-9130, 1588-2861},
	url = {https://link.springer.com/10.1007/s11192-025-05385-0},
	doi = {10.1007/s11192-025-05385-0},
	language = {en},
	number = {10},
	urldate = {2026-02-14},
	journal = {Scientometrics},
	author = {Cantone, Giulio Giacomo and Zheng, Er-Te and Tomaselli, Venera and Nightingale, Paul},
	month = oct,
	year = {2025},
	pages = {5345--5373},
}

@article{varnum_large_2024,
	title = {Large {Language} {Models} based on historical text could offer informative tools for behavioral science},
	volume = {121},
	issn = {0027-8424, 1091-6490},
	url = {https://pnas.org/doi/10.1073/pnas.2407639121},
	doi = {10.1073/pnas.2407639121},
	language = {en},
	number = {42},
	urldate = {2026-02-13},
	journal = {Proceedings of the National Academy of Sciences},
	author = {Varnum, Michael E. W. and Baumard, Nicolas and Atari, Mohammad and Gray, Kurt},
	month = oct,
	year = {2024},
	pages = {e2407639121},
}

@article{cave_hopes_2019,
	title = {Hopes and fears for intelligent machines in fiction and reality},
	volume = {1},
	issn = {2522-5839},
	url = {https://www.nature.com/articles/s42256-019-0020-9},
	doi = {10.1038/s42256-019-0020-9},
	language = {en},
	number = {2},
	urldate = {2023-10-09},
	journal = {Nature Machine Intelligence},
	author = {Cave, Stephen and Dihal, Kanta},
	month = feb,
	year = {2019},
	pages = {74--78},
}

@article{de_varda_cost_2025,
	title = {The cost of thinking is similar between large reasoning models and humans},
	volume = {122},
	issn = {0027-8424, 1091-6490},
	url = {https://pnas.org/doi/10.1073/pnas.2520077122},
	doi = {10.1073/pnas.2520077122},
	language = {en},
	number = {47},
	urldate = {2026-02-11},
	journal = {Proceedings of the National Academy of Sciences of the United States of America},
	author = {De Varda, Andrea Gregor and D’Elia, Ferdinando Pio and Kean, Hope and Lampinen, Andrew and Fedorenko, Evelina},
	month = nov,
	year = {2025},
	pages = {e2520077122},
}

@article{lampinen_language_2024,
	title = {Language models, like humans, show content effects on reasoning tasks},
	volume = {3},
	copyright = {https://creativecommons.org/licenses/by/4.0/},
	issn = {2752-6542},
	url = {https://academic.oup.com/pnasnexus/article/doi/10.1093/pnasnexus/pgae233/7712372},
	doi = {10.1093/pnasnexus/pgae233},
	language = {en},
	number = {7},
	urldate = {2026-02-11},
	journal = {PNAS Nexus},
	author = {Lampinen, Andrew K and Dasgupta, Ishita and Chan, Stephanie C Y and Sheahan, Hannah R and Creswell, Antonia and Kumaran, Dharshan and McClelland, James L and Hill, Felix},
	editor = {Abbott, Derek},
	month = jun,
	year = {2024},
	pages = {pgae233},
}

@article{tuckute_language_2024,
	title = {Language in brains, minds, and machines},
	issn = {0147-006X, 1545-4126},
	url = {https://www.annualreviews.org/content/journals/10.1146/annurev-neuro-120623-101142},
	doi = {10.1146/annurev-neuro-120623-101142},
	language = {en},
	urldate = {2024-05-13},
	journal = {Annual Review of Neuroscience},
	author = {Tuckute, Greta and Kanwisher, Nancy and Fedorenko, Evelina},
	month = apr,
    volume = {42},
    pages = {277--301},
	year = {2024},
}

@article{schrimpf_neural_2021,
	title = {The neural architecture of language: {Integrative} modeling converges on predictive processing},
	volume = {118},
	issn = {0027-8424, 1091-6490},
	shorttitle = {The neural architecture of language},
	url = {https://pnas.org/doi/full/10.1073/pnas.2105646118},
	doi = {10.1073/pnas.2105646118},
	language = {en},
	number = {45},
	urldate = {2026-02-11},
	journal = {Proceedings of the National Academy of Sciences of the United States of America},
	author = {Schrimpf, Martin and Blank, Idan Asher and Tuckute, Greta and Kauf, Carina and Hosseini, Eghbal A. and Kanwisher, Nancy and Tenenbaum, Joshua B. and Fedorenko, Evelina},
	month = nov,
	year = {2021},
	pages = {e2105646118},
}

@article{ferdman_ai_2025,
	title = {{AI} deskilling is a structural problem},
	issn = {0951-5666, 1435-5655},
	url = {https://link.springer.com/10.1007/s00146-025-02686-z},
	doi = {10.1007/s00146-025-02686-z},
	language = {en},
	urldate = {2026-02-03},
	journal = {AI \& Society},
	author = {Ferdman, Avigail},
	month = nov,
	year = {2025},
}

@article{yarkoni_generalizability_2022,
	title = {The generalizability crisis},
	volume = {45},
	issn = {0140-525X, 1469-1825},
	url = {https://www.cambridge.org/core/product/identifier/S0140525X20001685/type/journal_article},
	doi = {10.1017/S0140525X20001685},
	language = {en},
	urldate = {2026-01-27},
	journal = {Behavioral and Brain Sciences},
	author = {Yarkoni, Tal},
	year = {2022},
	pages = {e1},
}

@article{crielaard_refining_2024,
	title = {Refining the causal loop diagram: {A} tutorial for maximizing the contribution of domain expertise in computational system dynamics modeling.},
	volume = {29},
	copyright = {http://www.apa.org/pubs/journals/resources/open-access.aspx},
	issn = {1939-1463, 1082-989X},
	shorttitle = {Refining the causal loop diagram},
	url = {https://doi.apa.org/doi/10.1037/met0000484},
	doi = {10.1037/met0000484},
	language = {en},
	number = {1},
	urldate = {2026-01-19},
	journal = {Psychological Methods},
	author = {Crielaard, Loes and Uleman, Jeroen F. and Châtel, Bas D. L. and Epskamp, Sacha and Sloot, Peter M. A. and Quax, Rick},
	month = feb,
	year = {2024},
	pages = {169--201},
}

@article{mcgrath_2024,
  title   = {How can deep neural networks inform theory in psychological science?},
  author  = {McGrath, Sam Whitman and Russin, Jacob and Pavlick, Ellie and Feiman, Roman},
  journal = {Current Directions in Psychological Science},
  year    = {2024},
  volume  = {33},
  number  = {5},
  pages   = {325--333},
  language = {en},
  doi = {10.1177/09637214241268098},
}

@article{chambers_past_2021,
	title = {The past, present and future of {Registered} {Reports}},
	volume = {6},
	issn = {2397-3374},
	url = {https://www.nature.com/articles/s41562-021-01193-7},
	doi = {10.1038/s41562-021-01193-7},
	language = {en},
	number = {1},
	urldate = {2026-01-12},
	journal = {Nature Human Behaviour},
	author = {Chambers, Christopher D. and Tzavella, Loukia},
	month = nov,
	year = {2021},
	pages = {29--42},
}

@article{leising_tentative_2024,
	title = {A tentative roadmap for consensus building processes},
	volume = {5},
	issn = {2700-0710, 2700-0710},
	url = {https://journals.sagepub.com/doi/10.1177/27000710241298610},
	doi = {10.1177/27000710241298610},
	language = {en},
	urldate = {2025-02-09},
	journal = {Personality Science},
	author = {Leising, Daniel and Liesefeld, Heinrich and Buecker, Susanne and Glöckner, Andreas and Lortsch, Stefanie},
	month = apr,
	year = {2024},
	pages = {27000710241298610},
}

@article{Henrich2010,
	author = {Henrich, Joseph and Heine, Steven J. and Norenzayan, Ara},
	title = {The weirdest people in the world?},
	year = {2010},
	journal = {Behavioral and Brain Sciences},
	volume = {33},
	number = {2-3},
	pages = {61--83},
	doi = {10.1017/S0140525X0999152X},
}

@article{nishikawa2024llm_citation,
  title={Exploring the applicability of large language models to citation context analysis},
  author={Nishikawa, K. and Koshiba, H.},
  journal={Scientometrics},
  volume={129},
  pages={6751--6777},
  year={2024},
  doi={10.1007/s11192-024-05142-9}
}

@book{rumelhart86, editor = {Rumelhart, David E. and McClelland, James L. and PDP Research Group, CORPORATE}, title = {Parallel distributed processing: Explorations in the microstructure of cognition, vol. 1: Foundations}, year = {1986}, isbn = {026268053X}, publisher = {MIT Press}, address = {Cambridge, MA, USA}, url = {https://doi.org/10.7551/mitpress/5236.001.0001} }

@article{crockett_ai_2025,
	title = {{AI} surrogates and illusions of generalizability in cognitive science},
	issn = {13646613},
	url = {https://linkinghub.elsevier.com/retrieve/pii/S1364661325002517},
	doi = {10.1016/j.tics.2025.09.012},
	language = {en},
	urldate = {2025-10-25},
	journal = {Trends in Cognitive Sciences},
	author = {Crockett, M.J. and Messeri, Lisa},
	year = {2025},
	pages = {S1364661325002517},
}

@inproceedings{liesenfeld2024rethinking,
  title={Rethinking open source generative AI: Open washing and the EU AI Act},
  author={Liesenfeld, Andreas and Dingemanse, Mark},
  booktitle={Proceedings of the 2024 ACM Conference on Fairness, Accountability, and Transparency},
  pages={1774--1787},
  year={2024},
publisher = {Association for Computing Machinery},
address = {New York, USA},
doi={10.1145/3630106.3659005},
url={https://doi.10.1145/3630106.3659005}
}

@incollection{Read_Monroe_2023,
  author    = {Read, Stephen J. and Monroe, Brian M.},
  title     = {Computational models in personality and social psychology},
  booktitle = {The Cambridge handbook of computational cognitive sciences},
  editor    = {Sun, Ron},
  year      = {2023},
  pages     = {795--835},
  publisher = {Cambridge University Press},
  address   = {Cambridge, UK},
  collection = {Cambridge Handbooks in Psychology},
  url = {https://doi.org/10.1017/9781108755610.029}
}

@inproceedings{zhu_using_2025,
	title = {Using reinforcement learning to train large language models to explain human decisions},
	url = {http://arxiv.org/abs/2505.11614},
	doi = {10.48550/arXiv.2505.11614},
	language = {en},
	urldate = {2025-10-14},
	 booktitle = {The Fourteenth International Conference on Learning Representations},
  pages = {},
  address = {Rio de Janeiro, Brazil},
publisher = {ICLR},
	author = {Zhu, Jian-Qiao and Xie, Hanbo and Arumugam, Dilip and Wilson, Robert C. and Griffiths, Thomas L.},
	year = {2026},
}

@article{kieval_captured_2025,
	title = {“{Captured}” by centaur: {Opaque} predictions or process insights?},
	issn = {2329-8464, 2329-8456},
	shorttitle = {“{Captured}” by centaur},
	url = {https://doi.apa.org/doi/10.1037/xan0000410},
	doi = {10.1037/xan0000410},
	language = {en},
	urldate = {2025-10-13},
	journal = {Journal of Experimental Psychology: Animal Learning and Cognition},
	author = {Kieval, Phillip H. and Buckner, Cameron},
	month = sep,
	year = {2025},
}

@article{frank_cognitive_2025,
	title = {Cognitive modeling using artificial intelligence},
journal = {Annual Review of Psychology},
	language = {en},
	author = {Frank, Michael C. and Goodman, Noah D.},
    volume = {77},
    pages = {543--566},
	year = {2026},
    url = {https://doi.org/10.1146/annurev-psych-030625-040748},
    doi = {10.1146/annurev-psych-030625-040748}
}

@article{gentner_cognitive_2019,
	title = {Cognitive science is and should be pluralistic},
	volume = {11},
	issn = {1756-8757, 1756-8765},
	url = {https://onlinelibrary.wiley.com/doi/10.1111/tops.12459},
	doi = {10.1111/tops.12459},
	language = {en},
	number = {4},
	urldate = {2025-10-13},
	journal = {Topics in Cognitive Science},
	author = {Gentner, Dedre},
	month = oct,
	year = {2019},
	pages = {884--891},
}

@article{green_digital_2016,
	title = {A digital future for the history of psychology?},
	volume = {19},
	issn = {1939-0610, 1093-4510},
	url = {https://doi.apa.org/doi/10.1037/hop0000012},
	doi = {10.1037/hop0000012},
	language = {en},
	number = {3},
	urldate = {2025-10-13},
	journal = {History of Psychology},
	author = {Green, Christopher D.},
	year = {2016},
	pages = {209--219},
}

@incollection{buijsman_epistemic_2024,
  author    = {Buijsman, Stefan and Dur{\'a}n, Juan M.},
  title     = {Epistemic implications of machine learning models in science},
  booktitle = {The Routledge handbook of philosophy of scientific modeling},
  editor    = {Knuuttila, Tarja and Carrillo, Natalia and Koskinen, Rami},
  publisher = {Routledge},
  address   = {London, UK},
  year      = {2024},
  edition   = {1},
  chapter   = {33},
pages={456-468},
  doi       = {10.4324/9781003205647-39},
  language  = {en},
  urldate   = {2025-10-13},
  url = {https://doi.org/10.4324/9781003205647-39},
}

@article{binz_foundation_2025,
	title = {A foundation model to predict and capture human cognition},
	volume = {644},
	issn = {0028-0836, 1476-4687},
	url = {https://www.nature.com/articles/s41586-025-09215-4},
	doi = {10.1038/s41586-025-09215-4},
	language = {en},
	number = {8078},
	urldate = {2025-10-13},
	journal = {Nature},
	author = {Binz, Marcel and Akata, Elif and Bethge, Matthias and Brändle, Franziska and Callaway, Fred and Coda-Forno, Julian and Dayan, Peter and Demircan, Can and Eckstein, Maria K. and Éltető, Noémi and Griffiths, Thomas L. and Haridi, Susanne and Jagadish, Akshay K. and Ji-An, Li and Kipnis, Alexander and Kumar, Sreejan and Ludwig, Tobias and Mathony, Marvin and Mattar, Marcelo and Modirshanechi, Alireza and Nath, Surabhi S. and Peterson, Joshua C. and Rmus, Milena and Russek, Evan M. and Saanum, Tankred and Schubert, Johannes A. and Schulze Buschoff, Luca M. and Singhi, Nishad and Sui, Xin and Thalmann, Mirko and Theis, Fabian J. and Truong, Vuong and Udandarao, Vishaal and Voudouris, Konstantinos and Wilson, Robert and Witte, Kristin and Wu, Shuchen and Wulff, Dirk U. and Xiong, Huadong and Schulz, Eric},
	year = {2025},
	pages = {1002--1009},
}

@inproceedings{babaei_giglou_llms4synthesis_2024,
	address = {Hong Kong China},
	title = {{LLMs4S}ynthesis: {L}everaging large language models for scientific synthesis},
	isbn = {9798400710933},
	shorttitle = {{LLMs4Synthesis}},
	url = {https://dl.acm.org/doi/10.1145/3677389.3702565},
	doi = {10.1145/3677389.3702565},
	language = {en},
	urldate = {2025-10-13},
	booktitle = {Proceedings of the 24th {ACM}/{IEEE} {Joint} {Conference} on {Digital} {Libraries}},
	publisher = {ACM},
	author = {Babaei Giglou, Hamed and D'Souza, Jennifer and Auer, Sören},
	month = dec,
	year = {2024},
	pages = {1--12},
}

@article{frey2017risk,
  title = {Risk preference shares the psychometric structure of major psychological traits},
  author = {Frey, Renato and Pedroni, Andreas and Mata, Rui and Rieskamp, Jörg and Hertwig, Ralph},
  year = {2017},
  journal = {Scientific Advances},
  volume = {3},
  number = {10},
  pages = {e1701381},
  publisher = {American Association for the Advancement of Science},
  doi = {10.1126/sciadv.1701381},
  url = {https://www.science.org/doi/10.1126/sciadv.1701381},
  urldate = {2023-04-18}
}

@article{chalmers_what_2020,
	title = {What is conceptual engineering and what should it be?},
	issn = {0020-174X, 1502-3923},
	doi = {10.1080/0020174X.2020.1817141},
	language = {en},
	journal = {Inquiry},
	author = {Chalmers, David J.},
	year = {2020},
volume = {68},
number = {9},
	pages = {2902--2919},
}

@article{cummins_2025,
	title = {The threat of analytic flexibility in using large language models to simulate human data: {A} call to attention},
	language = {en},
journal={arXiv},
    year = {2025},
	author = {Cummins, Jamie},
doi={10.48550/arXiv.2509.13397},
}

@inproceedings{sen_2025,
	title = {Missing the margins: {A} systematic literature review on the demographic  representativeness of {LLMs}},
	language = {en},
	author = {Sen, Indira and Lutz, Marlene and Rogers, Elisa and Garcia, David and Strohmaier, Markus},
	editor = {Che, Wanxiang  and
      Nabende, Joyce  and
      Shutova, Ekaterina  and
      Pilehvar, Mohammad Taher},
booktitle = {Findings of the Association for Computational Linguistics: ACL 2025},
    month = {Jul},
    year = {2025},
    address = {Vienna, Austria},
    publisher = {Association for Computational Linguistics},
    url = {https://aclanthology.org/2025.findings-acl.1246/},
    doi = {10.18653/v1/2025.findings-acl.1246},
    pages = {24263--24289},
}

@article{LAIRD19871,
title = {{SOAR}: An architecture for general intelligence},
journal = {Artificial Intelligence},
volume = {33},
number = {1},
pages = {1-64},
year = {1987},
issn = {0004-3702},
doi = {10.1016/0004-3702(87)90050-6},
url = {https://www.sciencedirect.com/science/article/pii/0004370287900506},
author = {John E. Laird and Allen Newell and Paul S. Rosenbloom}
}

@incollection{sun_introduction_2001,
	edition = {1},
	title = {Introduction to computational cognitive modeling},
	copyright = {https://www.cambridge.org/core/terms},
	isbn = {978-0-511-81677-2 978-0-521-85741-3 978-0-521-67410-2},
	url = {https://www.cambridge.org/core/product/identifier/9780511816772%23c85741-ch1/type/book_part},
	language = {en},
	urldate = {2025-07-04},
	booktitle = {The Cambridge handbook of computational psychology},
	publisher = {Cambridge University Press},
address = {Cambridge, UK},
	author = {Sun, Ron},
	editor = {Sun, Ron},
	month = jan,
	year = {2001},
pages = {3--19},
	doi = {10.1017/CBO9780511816772.003},
	pages = {3--20},
}

@article{anderson_integrated_2004,
	title = {An integrated theory of the mind},
	volume = {111},
	issn = {1939-1471, 0033-295X},
	url = {https://doi.apa.org/doi/10.1037/0033-295X.111.4.1036},
	doi = {10.1037/0033-295X.111.4.1036},
	language = {en},
	number = {4},
	urldate = {2025-07-04},
	journal = {Psychological Review},
	author = {Anderson, John R. and Bothell, Daniel and Byrne, Michael D. and Douglass, Scott and Lebiere, Christian and Qin, Yulin},
	year = {2004},
	pages = {1036--1060}
}

@article{byrne_unified_2012,
	title = {Unified theories of cognition},
	volume = {3},
	copyright = {http://onlinelibrary.wiley.com/termsAndConditions\#vor},
	issn = {1939-5078, 1939-5086},
	url = {https://wires.onlinelibrary.wiley.com/doi/10.1002/wcs.1180},
	doi = {10.1002/wcs.1180},
	number = {4},
	urldate = {2025-07-04},
	journal = {WIREs Cognitive Science},
	author = {Byrne, Michael D.},
	year = {2012},
	pages = {431--438}
}

@article{wang_large_2025,
	title = {Large language models that replace human participants can harmfully misportray and flatten identity groups},
	volume = {7},
	issn = {2522-5839},
	url = {https://www.nature.com/articles/s42256-025-00986-z},
	doi = {10.1038/s42256-025-00986-z},
	language = {en},
	number = {3},
	urldate = {2025-07-04},
	journal = {Nature Machine Intelligence},
	author = {Wang, Angelina and Morgenstern, Jamie and Dickerson, John P.},
	year = {2025},
	pages = {400--411},
}

@book{snow1959two,
  author    = {Snow, C. P.},
  title     = {The two cultures and the scientific revolution},
  year      = {1959},
  publisher = {Cambridge University Press},
  address   = {New York, USA},
}

@article{michie_human_2017,
	title = {The {Human} {Behaviour}-{Change} {Project}: {H}arnessing the power of artificial intelligence and machine learning for evidence synthesis and interpretation},
	volume = {12},
	issn = {1748-5908},
	doi = {10.1186/s13012-017-0641-5},
	language = {en},
	number = {1},
	urldate = {2025-07-02},
	journal = {Implementation Science},
	author = {Michie, Susan and Thomas, James and Johnston, Marie and Aonghusa, Pol Mac and Shawe-Taylor, John and Kelly, Michael P. and Deleris, Léa A. and Finnerty, Ailbhe N. and Marques, Marta M. and Norris, Emma and O’Mara-Eves, Alison and West, Robert},
	year = {2017},
	pages = {121},
}

@article{elson_psychological_2023,
	title = {Psychological measures aren’t toothbrushes},
	volume = {1},
	issn = {2731-9121},
	doi = {10.1038/s44271-023-00026-9},
	language = {en},
	number = {1},
	urldate = {2024-04-25},
	journal = {Communications Psychology},
	author = {Elson, Malte and Hussey, Ian and Alsalti, Taym and Arslan, Ruben C.},
	year = {2023},
	pages = {25},
}

@article{van_rooij_formalizing_2020,
	title = {Formalizing verbal theories: {A} tutorial by dialogue},
	volume = {51},
	issn = {1864-9335, 2151-2590},
	shorttitle = {Formalizing {Verbal} {Theories}},
	url = {https://econtent.hogrefe.com/doi/10.1027/1864-9335/a000428},
	doi = {10.1027/1864-9335/a000428},
	language = {en},
	number = {5},
	journal = {Social Psychology},
	author = {Van Rooij, Iris and Blokpoel, Mark},
	year = {2020},
	pages = {285--298}
}

@article{peterson_using_2021,
	title = {Using large-scale experiments and machine learning to discover theories of human decision-making},
	volume = {372},
	issn = {0036-8075, 1095-9203},
	url = {https://www.sciencemag.org/lookup/doi/10.1126/science.abe2629},
	doi = {10.1126/science.abe2629},
	language = {en},
	number = {6547},
	urldate = {2021-06-11},
	journal = {Science},
	author = {Peterson, Joshua C. and Bourgin, David D. and Agrawal, Mayank and Reichman, Daniel and Griffiths, Thomas L.},
	year = {2021},
	pages = {1209--1214}
}

@article{meehl_1990,
	title = {Why summaries of research on psychological theories are often uninterpretable},
	language = {en},
	author = {Meehl, Paul E},
journal = {Psychological Reports},
volume = {66},
number = {1},
pages = {195-244},
year = {1990},
doi = {10.2466/PR0.66.1.195-244}
}

@article{wulff_mata_2025, 
    title={Escaping the jingle–jangle jungle: Increasing conceptual clarity in psychology using large language models}, 
    url={osf.io/preprints/psyarxiv/ksuh8_v1}, 
    DOI={10.1177/09637214251382083}, 
    publisher={PsyArXiv}, 
    author={Wulff, Dirk U and Mata, Rui}, 
volume = {},
number = {},
pages = {09637214251382083},
journal = {Current Directions in Psychological Science},
    year={2025}}

@article{norris_scoping_2019,
	title = {A scoping review of ontologies related to human behaviour change},
	volume = {3},
	issn = {2397-3374},
	url = {http://www.nature.com/articles/s41562-018-0511-4},
	doi = {10.1038/s41562-018-0511-4},
	language = {en},
	number = {2},
	urldate = {2022-05-23},
	journal = {Nature Human Behaviour},
	author = {Norris, Emma and Finnerty, Ailbhe N. and Hastings, Janna and Stokes, Gillian and Michie, Susan},
	year = {2019},
	pages = {164--172},
}

@article{sharp_use_2023,
	title = {The use of ontologies to accelerate the behavioral sciences: Promises and challenges},
	issn = {0963-7214, 1467-8721},
	doi = {10.1177/09637214231183917},
	language = {en},
	journal = {Current Directions in Psychological Science},
	author = {Sharp, Carla and Kaplan, Robert M. and Strauman, Timothy J.},
	year = {2023},
volume = {32},
number = {5},
	pages = {418--426},
}

@book{jacobs2014defense,
  title={In defense of disciplines: Interdisciplinarity and specialization in the research university},
  author={Jacobs, Jerry A},
  year={2014},
  publisher={University of Chicago Press},
address = {Chicago, IL, USA},
doi = {10.7208/chicago/9780226069463.001.0001},
url = {https://doi.10.7208/chicago/9780226069463.001.0001}
}

@book{gardner_minds_1985,
	address = {New York,  USA},
	title = {The mind's new science: {A} history of the cognitive revolution.},
	publisher = {Basic Books},
	author = {Gardner, Howard},
	year = {1985},
}

@article{nunez2019happened,
  title={What happened to cognitive science?},
  author={N{\'u}{\~n}ez, Rafael and Allen, Michael and Gao, Richard and Miller Rigoli, Carson and Relaford-Doyle, Josephine and Semenuks, Arturs},
  journal={Nature Human Behaviour},
  volume={3},
  number={8},
  pages={782--791},
  year={2019},
  DOI={10.1038/s41562-019-0626-2},
publisher={Nature Publishing Group UK London}
}

@article{eronen2021theory,
  title={The theory crisis in psychology: How to move forward},
  author={Eronen, Markus I and Bringmann, Laura F},
  journal={Perspectives on Psychological Science},
  volume={16},
  number={4},
  pages={779--788},
  year={2021},
  publisher={Sage Publications Sage CA: Los Angeles, CA},
doi={10.1177/1745691620970586},
}

@article{van2024reclaiming,
  title={Reclaiming {AI} as a theoretical tool for cognitive science},
  author={Van Rooij, Iris and Guest, Olivia and Adolfi, Federico and de Haan, Ronald and Kolokolova, Antonina and Rich, Patricia},
  journal={Computational Brain \& Behavior},
  volume={7},
  number={4},
  pages={616--636},
  year={2024},
doi={10.1007/s42113-024-00217-5},
  publisher={Springer}
}

@article{taher2023embedding,
  title={An embedding approach for analyzing the evolution of research topics with a case study on computer science subdomains},
  author={Taher Harikandeh, Seyyed Reza and Aliakbary, Sadegh and Taheri, Soroush},
  journal={Scientometrics},
  volume={128},
  number={3},
  pages={1567--1582},
  year={2023},
  publisher={Springer},
doi={10.1007/s11192-023-04642-4}
}

@article{anvari_defragmenting_2025,
	title = {Defragmenting psychology},
	issn = {2397-3374},
	url = {https://www.nature.com/articles/s41562-025-02138-0},
	doi = {10.1038/s41562-025-02138-0},
	language = {en},
	urldate = {2025-03-29},
	journal = {Nature Human Behaviour},
volume =  {9},
pages = {836--839},
	author = {Anvari, Farid and Alsalti, Taym and Oehler, Lorenz A. and Hussey, Ian and Elson, Malte and Arslan, Ruben C.},
	year = {2025},
}

@inproceedings{babaei2023llms4ol,
  title={LLMs4OL: Large language models for ontology learning},
  author={Babaei Giglou, Hamed and D’Souza, Jennifer and Auer, S{\"o}ren},
  booktitle={The Semantic Web---ISWC 2023},
  pages={408--427},
  year={2023},
publisher = {Springer Nature},
address = {Cham, Switzerland},
url={https://doi.10.1007/978-3-031-47240-4_22}
}

@article{larsen2016tool,
  title={A tool for addressing construct identity in literature reviews and meta-analyses},
  author={Larsen, Kai R and Bong, Chih How},
  journal={Mis Quarterly},
  volume={40},
  number={3},
  pages={529--552},
  year={2016},
  publisher={JSTOR},
doi={10.25300/misq/2016/40.3.01},
}

@article{wulff2025semantic,
  title={Semantic embeddings reveal and address taxonomic incommensurability in psychological measurement},
  author={Wulff, Dirk U and Mata, Rui},
  journal={Nature Human Behaviour},
volume={9},
number={5},
  pages={944–954},
  year={2025},
DOI={10.1038/s41562-024-02089-y},
  publisher={Nature Publishing Group UK London}
}

@article{hommel2024language,
  title={Language models accurately infer correlations between psychological items and scales from text alone},
  author={Hommel, Bj{\"o}rn E and Arslan, Ruben C},
    doi = {10.1177/25152459251377093},
  year={2025},
volume={8},
number={4},
  journal={Advances in Methods and Practices in Psychological Science}}

@article{rosenbusch2020semantic,
  title={The Semantic Scale Network: An online tool to detect semantic overlap of psychological scales and prevent scale redundancies.},
  author={Rosenbusch, Hannes and Wanders, Florian and Pit, Ilse L},
  journal={Psychological Methods},
  volume={25},
  number={3},
  pages={380},
  year={2020},
doi={10.1037/met0000244},
  publisher={American Psychological Association}
}

@article{musslick2025automating,
  title={Automating the practice of science: Opportunities, challenges, and implications},
  author={Musslick, Sebastian and Bartlett, Laura K and Chandramouli, Suyog H and Dubova, Marina and Gobet, Fernand and Griffiths, Thomas L and Hullman, Jessica and King, Ross D and Kutz, J Nathan and Lucas, Christopher G and others},
  journal={Proceedings of the National Academy of Sciences of the United States of America},
  volume={122},
  number={5},
  pages={e2401238121},
  year={2025},
  publisher={National Academy of Sciences},
doi={10.1073/pnas.2401238121},
}

@article{luo2024large,
  title={Large language models surpass human experts in predicting neuroscience results},
  author={Luo, Xiaoliang and Rechardt, Akilles and Sun, Guangzhi and Nejad, Kevin K and Y{\'a}{\~n}ez, Felipe and Yilmaz, Bati and Lee, Kangjoo and Cohen, Alexandra O and Borghesani, Valentina and Pashkov, Anton and others},
  journal={Nature Human Behaviour},
volume={9},
number={2},
  pages={305--315},
  year={2024},
doi={10.1038/s41562-024-02046-9},
  publisher={Nature Publishing Group UK London}
}

@article{narayanan2025overreliance,
  title={Why an overreliance on {AI}-driven modelling is bad for science},
  author={Narayanan, Arvind and Kapoor, Sayash},
  journal={Nature},
  volume={640},
  number={8058},
  pages={312--314},
  year={2025},
doi={10.1038/d41586-025-01067-2},
  publisher={Nature Publishing Group UK London}
}

@inproceedings{balloccu2024leak,
  title={Leak, cheat, repeat: Data contamination and evaluation malpractices in closed-source {LLM}s},
  author={Balloccu, Simone and Schmidtov{\'a}, Patr{\'\i}cia and Lango, Mateusz and Du{\v{s}}ek, Ond{\v{r}}ej},
 editor = {Graham, Yvette  and      Purver, Matthew},
pages = {67—93},
booktitle = {Proceedings of the 18th Conference of the European Chapter of the Association for Computational Linguistics (Volume 1: Long Papers)},
address = {St. Julian{'}s, Malta},
publisher = {Association for Computational Linguistics},
url={https://aclanthology.org/2024.eacl-long.5/},
  year={2024}
}

@article{wulff2024behavioral,
  title={The behavioral and social sciences need open LLMs},
  author={Wulff, Dirk U and Hussain, Zak and Mata, Rui},
  year={2024},
  journal={Open Science Framework},
doi={10.31219/osf.io/ybvzs}
}

@article{palmer2024using,
  title={Using proprietary language models in academic research requires explicit justification},
  author={Palmer, Alexis and Smith, Noah A and Spirling, Arthur},
  journal={Nature Computational Science},
  volume={4},
  number={1},
  pages={2--3},
doi={10.1038/s43588-023-00585-1},
  year={2024},
  publisher={Nature Publishing Group US New York}
}

@article{hussain2024tutorial,
  title={A tutorial on open-source large language models for behavioral science},
  author={Hussain, Zak and Binz, Marcel and Mata, Rui and Wulff, Dirk U},
  journal={Behavior Research Methods},
  volume={56},
  number={8},
  pages={8214--8237},
  year={2024},
  publisher={Springer},
doi={https://doi.org/10.31234/osf.io/f7stn}
}

@article{cronbach_1957,
	title = {The two disciplines of scientific psychology},
	language = {en},
	journal = {The American Psychologist},
	author = {Cronbach, Lee J},
    volume={12},
    number={11},
    pages={671--684},
    year={1957},
    DOI={https://doi.org/10.1037/h0043943}
}

@article{poldrack_brain_2016,
	title = {From brain maps to cognitive ontologies: {Informatics} and the search for mental structure},
	volume = {67},
	doi = {10.1146/annurev-psych-122414-033729},
	language = {en},
	number = {1},
	urldate = {2025-05-28},
	journal = {Annual Review of Psychology},
	author = {Poldrack, Russell A. and Yarkoni, Tal},
	year = {2016},
	pages = {587--612}
}

@article{rmus_generating_2025,
	title = {Generating computational cognitive models using large language models},
	url = {http://arxiv.org/abs/2502.00879},
	doi = {10.48550/arXiv.2502.00879},
	language = {en},
	urldate = {2025-05-25},
	journal = {arXiv},
	author = {Rmus, Milena and Jagadish, Akshay K. and Mathony, Marvin and Ludwig, Tobias and Schulz, Eric},
	year = {2025}
}

@article{waaijers_theoraizer_2024,
	title = {theoraizer: {AI}-assisted theory construction},
	url = {https://osf.io/gu9yq},
	doi = {10.31234/osf.io/gu9yq},
	language = {en},
	author = {Waaijers, Meike and Rosenbusch, Hannes and Van Lissa, Caspar J. and Roefs, Anne and Borsboom, Denny},
	month = aug,
	year = {2024},
    note={OSF},
    journal={PsyArXiv}}

@article{oberauer_addressing_2019,
	title = {Addressing the theory crisis in psychology},
	volume = {26},
	issn = {1069-9384, 1531-5320},
	doi = {10.3758/s13423-019-01645-2},
	language = {en},
	number = {5},
	journal = {Psychonomic Bulletin \& Review},
	author = {Oberauer, Klaus and Lewandowsky, Stephan},
	year = {2019},
	pages = {1596--1618}
}

@article{sanbonmatsu_there_2025,
	title = {There is no theory crisis in psychological science.},
	doi = {10.1037/teo0000301},
	language = {en},
	journal = {Journal of Theoretical and Philosophical Psychology},
	author = {Sanbonmatsu, David M. and Neufeld, Becky and Posavac, Steven S.},
	year = {2025},
}

@article{smaldino_how_2020,
	title = {How to translate a verbal theory into a formal model},
	volume = {51},
	doi = {10.1027/1864-9335/a000425},
	language = {en},
	number = {4},
	journal = {Social Psychology},
	author = {Smaldino, Paul E.},
	year = {2020},
	pages = {207--218}
}

@inproceedings{newell_you_1973,
	title = {You can’t play 20 questions with nature and win: Projective comments on the papers of this symposium},
	language = {en},
	urldate = {2022-02-28},
	booktitle = {Visual {Information} {Processing}},
	publisher = {Academic Press (Elsevier)},
  address   = {New York, USA},
	author = {Newell, Allen},
editor = {Chase, William G.},
	year = {1973},
	doi = {https://doi.org/10.1016/B978-0-12-170150-5.50012-3},
	pages = {283--308}
}

@article{yarkoni_choosing_2017,
	title = {Choosing prediction over explanation in psychology: {Lessons} from machine learning},
	volume = {12},
	url = {https://journals.sagepub.com/doi/10.1177/1745691617693393},
	doi = {10.1177/1745691617693393},
	language = {en},
	number = {6},
	journal = {Perspectives on Psychological Science},
	author = {Yarkoni, Tal and Westfall, Jacob},
	year = {2017},
	pages = {1100--1122}
}

@article{savcisens_using_2023,
	title = {Using sequences of life-events to predict human lives},
	volume = {4},
	issn = {2662-8457},
	url = {https://www.nature.com/articles/s43588-023-00573-5},
	doi = {10.1038/s43588-023-00573-5},
	language = {en},
	number = {1},
	journal = {Nature Computational Science},
	author = {Savcisens, Germans and Eliassi-Rad, Tina and Hansen, Lars Kai and Mortensen, Laust Hvas and Lilleholt, Lau and Rogers, Anna and Zettler, Ingo and Lehmann, Sune},
	year = {2023},
	pages = {43--56}
}

@inproceedings{havaldar_building_2024,
	    title = {Building knowledge-guided lexica to model cultural variation},
    author = {Havaldar, Shreya  and
      Giorgi, Salvatore  and
      Rai, Sunny  and
      Talhelm, Thomas  and
      Guntuku, Sharath Chandra  and
      Ungar, Lyle},
    editor = {Duh, Kevin  and
      Gomez, Helena  and
      Bethard, Steven},
    booktitle = {Proceedings of the 2024 Conference of the North American Chapter of the Association for Computational Linguistics: Human Language Technologies (Volume 1: Long Papers)},
    month = {jun},
    year = {2024},
    address = {Mexico City, Mexico},
    publisher = {Association for Computational Linguistics},
url = {https://doi.org/10.18653/v1/2024.naacl-long.12},
    pages = {211-226},

}

@article{barrett_towards_2020,
	title = {Towards a cognitive science of the human: {Cross}-cultural approaches and their urgency},
	volume = {24},
	issn = {13646613},
	shorttitle = {Towards a {Cognitive} {Science} of the {Human}},
	url = {https://linkinghub.elsevier.com/retrieve/pii/S1364661320301315},
	doi = {10.1016/j.tics.2020.05.007},
	language = {en},
	number = {8},
	journal = {Trends in Cognitive Sciences},
	author = {Barrett, H. Clark},
	year = {2020},
	pages = {620--638}
}

@article{wig_participant_2024,
	title = {Participant diversity is necessary to advance brain aging research},
	volume = {28},
	issn = {13646613},
	url = {https://linkinghub.elsevier.com/retrieve/pii/S1364661323003017},
	doi = {10.1016/j.tics.2023.12.004},
	language = {en},
	number = {2},
	journal = {Trends in Cognitive Sciences},
	author = {Wig, Gagan S. and Klausner, Sarah and Chan, Micaela Y. and Sullins, Cameron and Rayanki, Anirudh and Seale, Maya},
	year = {2024},
	pages = {92--96}
}

@article{danziger_psychology_2013,
	title = {Psychology and its history},
	volume = {23},
	issn = {0959-3543, 1461-7447},
    doi = {10.1177/0959354313502746},
	language = {en},
	number = {6},
	journal = {Theory \& Psychology},
	author = {Danziger, Kurt},
	year = {2013},
	pages = {829--839},
}

@article{hochstein_categorizing_2016,
	title = {Categorizing the mental},
	volume = {66},
	issn = {0031-8094, 1467-9213},
	doi = {10.1093/pq/pqw001},
	language = {en},
	number = {265},
	journal = {The Philosophical Quarterly},
	author = {Hochstein, Eric},
	year = {2016},
	pages = {745-759},
}

@article{hertwig2019three,
  title={Three gaps and what they may mean for risk preference},
  author={Hertwig, Ralph and Wulff, Dirk U and Mata, Rui},
  journal={Philosophical Transactions of the Royal Society Series B, Biological Sciences},
  volume={374},
  number={1766},
  pages={20180140},
  year={2019},
doi={10.1098/rstb.2018.0140},
  publisher={The Royal Society},
}

@article{thoma2025mapping,
  author       = {Thoma, Anna I. and Bolenz, Florian and Tiede, Kevin and Yang, Yujia and Palminteri, Stefano and Hertwig, Ralph and Wulff, Dirk U.},
  title        = {Mapping the landscape of behavioral reinforcement learning research},
  year         = {2025},
  doi          = {10.31234/osf.io/6c2va_v1},
  journal         = {PsyArXiv},
}

@article{fulawka2025reasons,
  author       = {Fulawka, Kamil and Hertwig, Ralph and Wulff, Dirk U.},
  title        = {Large language models accurately identify decision reasons in verbal reports},
  year         = {2025},
  doi          = {10.31234/osf.io/yuzmw_v1},
  howpublished = {\url{https://doi.org/10.31234/osf.io/yuzmw_v1}},
  journal         = {PsyArXiv},
}

@article{van2025combining,
  title={Combining psychology with artificial intelligence: What could possibly go wrong?},
  author={{van Rooij}, Iris and Guest, Olivia},
  year={2025},
  doi          = {10.31234/osf.io/aue4m_v1},
  journal         = {PsyArXiv},
}

@article{guest2021computational,
  title={How computational modeling can force theory building in psychological science},
  author={Guest, Olivia and Martin, Andrea E},
  journal={Perspectives on Psychological Science},
  volume={16},
  number={4},
  pages={789--802},
  year={2021},
doi={10.1177/1745691620970585},
  publisher={Sage Publications Sage CA: Los Angeles, CA}
}

@article{xie2025centaur,
  title={Centaur may have learned a shortcut that explains away psychological tasks},
  author={Xie, Hanbo and Zhu, Jian-Qiao},
  year={2025},
  doi          = {10.31234/osf.io/u7z4t_v2},
  howpublished = {\url{https://doi.org/10.31234/osf.io/u7z4t_v2}},
  journal         = {PsyArXiv},
}

@article{schroder2025large,
  title={Large language models do not simulate human psychology},
  author={Schr{\"o}der, Sarah and Morgenroth, Thekla and Kuhl, Ulrike and Vaquet, Valerie and Paa{\ss}en, Benjamin},
  year={2025},
  doi          = {10.48550/arXiv.2508.06950},
  howpublished = {\url{https://doi.org/10.48550/arXiv.2508.06950}},
  Journal         = {arXiv},
}

@inproceedings{demircan2024sparse,
  title={Sparse autoencoders reveal temporal difference learning in large language models},
  author={Demircan, Can and Saanum, Tankred and Jagadish, Akshay K and Binz, Marcel and Schulz, Eric},
  booktitle = {The Thirteenth International Conference on Learning Representations},
  year={2025},
  pages = {4972},
  address = {Appleton, WI, USA},
publisher = {ICLR},
  url          = {https://openreview.net/forum?id=2tIyA5cri8},
}

@article{hussain2024probing,
  title={Probing the contents of semantic representations from text, behavior, and brain data using the psychNorms metabase},
  author={Hussain, Zak and Mata, Rui and Newell, Ben R and Wulff, Dirk U},
  year={2024},
  doi          = {10.48550/arXiv.2412.04936},
  howpublished = {\url{https://doi.org/10.48550/arXiv.2412.04936}},
  journal         = {arXiv},
}

@inproceedings{haller2023opiniongpt,
  title={Opinion{GPT}: Modelling explicit biases in instruction-tuned {LLMs}},
  author={Haller, Patrick and Aynetdinov, Ansar and Akbik, Alan},
  year={2024},
    editor = "Chang, Kai-Wei  and
      Lee, Annie  and
      Rajani, Nazneen",
    booktitle = {Proceedings of the 2024 Conference of the North American Chapter of the Association for Computational Linguistics: Human Language Technologies (Volume 3: System Demonstrations)},
    month = {Jun},
    year = {2024},
    address = {Mexico City, Mexico},
    publisher = {Association for Computational Linguistics},
    url = {https://aclanthology.org/2024.naacl-demo.8/},
    doi = {10.18653/v1/2024.naacl-demo.8},
    pages = {78--86},
}

@inproceedings{suh2025language,
  title={Language model fine-tuning on scaled survey data for predicting distributions of public opinions},
  author={Suh, Joseph and Jahanparast, Erfan and Moon, Suhong and Kang, Minwoo and Chang, Serina},
      editor = {Che, Wanxiang  and
      Nabende, Joyce  and
      Shutova, Ekaterina  and
      Pilehvar, Mohammad Taher},
    booktitle = {Proceedings of the 63rd Annual Meeting of the Association for Computational Linguistics (Volume 1: Long Papers)},
    month = {Jul},
    year = {2025},
    address = {Vienna, Austria},
    publisher = {Association for Computational Linguistics},
    doi = {10.18653/v1/2025.acl-long.1028},
    url={https://doi.10.18653/v1/2025.acl-long.1028},
    pages = {21147--21170},
}

@inproceedings{tosato2025persistent,
  title={Persistent instability in LLM's personality measurements: Effects of scale, reasoning, and conversation history},
  author={Tosato, Tommaso and Helbling, Saskia and Mantilla-Ramos, Yorguin-Jose and Hegazy, Mahmood and Tosato, Alberto and Lemay, David John and Rish, Irina and Dumas, Guillaume},
booktitle = {Proceedings of the 40th AAAI Conference on Artificial Intelligence},
month  = {Jan},
year = {2026},
address = {Washington, DC, USA},
publisher = {AAAI Press},
doi={10.48550/arXiv.2509.13397},
url={https://doi.org/10.48550/arXiv.2509.13397},
pages = {00--00},
}

@article{bhatia2025computational,
  title={Computational analysis of 100 K choice dilemmas: Decision attributes, trade-off structures, and model-based prediction},
  author={Bhatia, Sudeep and van Baal, Simon T and Wang, Feiyi and Walasek, Lukasz},
  journal={Proceedings of the National Academy of Sciences of the United States of America},
  volume={122},
  number={17},
  pages={e2406489122},
  year={2025},
  publisher={National Academy of Sciences},
doi={10.1073/pnas.2406489122},
}

@article{barrie2025emergent,
  title={Emergent {LLM} behaviors are observationally equivalent to data leakage},
  author={Barrie, Christopher and T{\"o}rnberg, Petter},
  journal={arXiv},
  year={2025},
doi={10.48550/arXiv.2505.23796},
}

@article{ashery2025reply,
  title={Reply to "{E}mergent {LLM} behaviors are observationally equivalent to data leakage"},
  author={Ashery, Ariel Flint and Aiello, Luca Maria and Baronchelli, Andrea},
  journal={arXiv},
  year={2025},
doi={10.48550/arXiv.2506.18600},
}

@InProceedings{sadruddin2025llms4schemadiscovery,
  title={LLMs4SchemaDiscovery: A human-in-the-loop workflow for scientific schema mining with large language models},
  author={Sadruddin, Sameer and D’Souza, Jennifer and Poupaki, Eleni and Watkins, Alex and Babaei Giglou, Hamed and Rula, Anisa and Karasulu, Bora and Auer, S{\"o}ren and Mackus, Adrie and Kessels, Erwin},
  booktitle={The Semantic Web},
  pages={244--261},
editor={Curry, Edward and Acosta, Maribel
and Poveda-Villal{\'o}n, Maria
and van Erp, Marieke
and Ojo, Adegboyega
and Hose, Katja
and Shimizu, Cogan
and Lisena, Pasquale
},
  year={2025},
  publisher={Springer Nature},
address={Cham, Switzerland},
url={https://doi.10.1007/978-3-031-94578-6_14},
}

@inproceedings{giglou2024llms4om,
  title={{LLMs4OM}: Matching ontologies with large language models},
  author={Babaei Giglou, Hamed and D'Souza, Jennifer and Engel, Felix and Auer, S{\"o}ren},
booktitle = {The Semantic Web: ESWC 2024 Satellite Events},
pages = {25--35},
publisher = {Springer Nature},
address = {Cham, Switzerland},
year={2025},
url={https://doi.10.1007/978-3-031-78952-6_3},
}

@article{shannon1951prediction,
  title={Prediction and entropy of printed English},
  author={Shannon, Claude E},
  journal={Bell System Technical Journal},
  volume={30},
  number={1},
  pages={50--64},
  year={1951},
  publisher={Wiley Online Library},
doi={10.1002/j.1538-7305.1951.tb01366.x},
}

@incollection{firth1957synopsis,
  title={A synopsis of linguistic theory, 1930-1955},
  author={Firth, John Rupert},
  booktitle={Studies in linguistic analysis. Special volume of the Philological Society},
  pages={10--32},
  year={1957},
  publisher={Blackwell},
  address={Oxford, UK}
}

@article{harris1954distributional,
  title={Distributional structure},
  author={Harris, Zellig S},
  journal={Word},
  volume={10},
  number={2-3},
  pages={146--162},
  year={1954},
  publisher={Taylor \& Francis},
  doi={10.1080/00437956.1954.11659520}
}

@inproceedings{vaswani2017attention,
 author = {Vaswani, Ashish and Shazeer, Noam and Parmar, Niki and Uszkoreit, Jakob and Jones, Llion and Gomez, Aidan N and Kaiser, \L ukasz and Polosukhin, Illia},
 booktitle = {Advances in Neural Information Processing Systems},
 editor = {I. Guyon and U. Von Luxburg and S. Bengio and H. Wallach and R. Fergus and S. Vishwanathan and R. Garnett},
 pages = {},
 publisher = {Curran Associates, Inc.},
 title = {Attention is all you need},
 url = {https://proceedings.neurips.cc/paper_files/paper/2017/file/3f5ee243547dee91fbd053c1c4a845aa-Paper.pdf},
 volume = {30},
 year = {2017},
}

@inproceedings{brown2020language,
 author = {Brown, Tom and Mann, Benjamin and Ryder, Nick and Subbiah, Melanie and Kaplan, Jared D and Dhariwal, Prafulla and Neelakantan, Arvind and Shyam, Pranav and Sastry, Girish and Askell, Amanda and Agarwal, Sandhini and Herbert-Voss, Ariel and Krueger, Gretchen and Henighan, Tom and Child, Rewon and Ramesh, Aditya and Ziegler, Daniel and Wu, Jeffrey and Winter, Clemens and Hesse, Chris and Chen, Mark and Sigler, Eric and Litwin, Mateusz and Gray, Scott and Chess, Benjamin and Clark, Jack and Berner, Christopher and McCandlish, Sam and Radford, Alec and Sutskever, Ilya and Amodei, Dario},
 booktitle = {Advances in Neural Information Processing Systems},
 editor = {H. Larochelle and M. Ranzato and R. Hadsell and M.F. Balcan and H. Lin},
 pages = {1877--1901},
 publisher = {Curran Associates, Inc.},
 title = {Language models are few-shot learners},
 url = {https://proceedings.neurips.cc/paper_files/paper/2020/file/1457c0d6bfcb4967418bfb8ac142f64a-Paper.pdf},
 volume = {33},
 year = {2020}
}

@inproceedings{devlin2019bert,
title = {{BERT}: Pre-training of deep bidirectional transformers for language understanding},
author = {Devlin, Jacob and Chang, Ming-Wei and Lee, Kenton and Toutanova, Kristina},
editor = {Burstein, Jill and Doran, Christy and Solorio, Thamar},
booktitle = {Proceedings of the 2019 Conference of the North {A}merican Chapter of the Association for Computational Linguistics: Human Language Technologies, Volume 1 (Long and Short Papers)},
month = Jun,
year = {2019},
address = {Minneapolis, Minnesota},
publisher = {Association for Computational Linguistics},
url = {https://aclanthology.org/N19-1423/},
doi = {10.18653/v1/N19-1423},
pages = {4171—4186},
}

@inproceedings{reimers2019sentence,
title = {Sentence-{BERT}: Sentence embeddings using {S}iamese {BERT}-networks},
author = "Reimers, Nils and Gurevych, Iryna",
editor = "Inui, Kentaro and Jiang, Jing and Ng, Vincent and Wan, Xiaojun",
booktitle = "Proceedings of the 2019 Conference on Empirical Methods in Natural Language Processing and the 9th International Joint Conference on Natural Language Processing (EMNLP-IJCNLP)",
month = nov,
year = "2019",
address = "Hong Kong, China",
publisher = "Association for Computational Linguistics",
url = "https://aclanthology.org/D19-1410/",
doi = "10.18653/v1/D19-1410",
pages = "3982--3992"
}

\end{document}